\documentclass[11pt]{article}
\usepackage{fullpage}
\usepackage[margin=1in]{geometry}

\usepackage{graphicx}
\usepackage{natbib}
\usepackage{xcolor}
\usepackage{hyperref}
\usepackage{algorithm,algpseudocode}
\usepackage{algorithmicx}
\usepackage{multicol}
\usepackage{multirow}
\usepackage{hhline}
\usepackage{amsmath}
\usepackage{mathtools}
\usepackage[capitalize,noabbrev]{cleveref}
\usepackage{float}
\usepackage{array}
\usepackage{xspace}
\usepackage{subfigure}
\newcolumntype{L}[1]{>{\raggedright\let\newline\\\arraybackslash\hspace{0pt}}m{#1}}
\newcolumntype{C}[1]{>{\centering\let\newline\\\arraybackslash\hspace{0pt}}m{#1}}
\newcolumntype{R}[1]{>{\raggedleft\let\newline\\\arraybackslash\hspace{0pt}}m{#1}}

\RequirePackage[T1]{fontenc} 
\RequirePackage[tt=false, type1=true]{libertine} 
\RequirePackage[varqu]{zi4} 
\RequirePackage[libertine]{newtxmath}

\newtheorem{thm}{Theorem}[section]
\newtheorem{lem}[thm]{Lemma}
\newtheorem{pro}[thm]{Proposition}
\newtheorem{cor}[thm]{Corollary}

\newtheorem{dfn}[thm]{Definition}

\newtheorem{ass}[thm]{Assumption}

\newcommand{\calt}{\mathcal{T}}

\newcommand{\cali}{\mathcal{I}}

\newcommand{\la}{{\lambda}}

\newcommand{\opt}{\texttt{OPT}\xspace}
\newcommand{\alg}{\texttt{ALG}\xspace}
\newcommand{\ota}{\texttt{OTA}\xspace}

\newcommand{\osid}{\texttt{OSID}\xspace}

\newcommand{\nostr}{\texttt{NoSTR}\xspace}
\newcommand{\arp}{\texttt{OS-on}\xspace}
\newcommand{\osbase}{\texttt{OS-b}\xspace}
\newcommand{\osalf}{\texttt{OS-}$\lambda$\xspace}
\newcommand{\arpia}{\texttt{OS-on-IA}\xspace}
\newcommand{\osalfia}{\texttt{OS}-$\lambda$\texttt{-IA}\xspace}

\newcommand{\osstatic}{\texttt{OS}-$\lambda^*$\xspace}
\newcommand{\otpalgwc}{\texttt{OTA-on}\xspace}
\newcommand{\otpalgoff}{\texttt{OTA}$(\lambda^{*})$\xspace}
\newcommand{\naivebaseline}{\texttt{base}$(\lambda^{\texttt{alf}})$\xspace}
\newcommand{\otpalgalf}{\texttt{OTA}$(\lambda^{\texttt{alf}})$\xspace}

\newcommand{\revision}[1]{{\textcolor{black}{#1}}}

\newcommand{\pmax}{p_{\text{max}}}
\newcommand{\pmin}{p_{\text{min}}}

\begin{document}

\title{Online Search with Predictions: Pareto-optimal Algorithm and its Applications in Energy Markets }

\def\thefootnote{$\#$}\footnotetext{Both authors contribute equally to this work.}\def\thefootnote{\arabic{footnote}}

\author{Russell Lee$^\#$\thanks{University of Massachusetts Amherst. Email: {\tt  rclee@cs.umass.edu}.} \and
Bo~Sun$^\#$\thanks{University of Waterloo. Email: {\tt bo.sun@uwaterloo.ca}.}
\and
Mohammad~Hajiesmaili\thanks{University of Massachusetts Amherst. Email: {\tt hajiesmaili@cs.umass.edu}.} 
\and 
John C.S. Lui\thanks{The Chinese University of Hong Kong. Email: {\tt cslui@cse.cuhk.edu.hk. }}
}

\begin{titlepage}
\maketitle

\thispagestyle{empty}

\begin{abstract}
This paper develops learning-augmented algorithms for energy trading in volatile electricity markets. 
The basic problem is to sell (or buy) $k$ units of energy for the highest revenue (lowest cost) over uncertain time-varying prices, which can framed as a classic online search problem in the literature of competitive analysis. 
State-of-the-art algorithms assume no knowledge about future market prices when they make trading decisions in each time slot, and aim for guaranteeing the performance for the worst-case price sequence.
In practice, however, predictions about future prices become commonly available by leveraging machine learning. 
This paper aims to incorporate machine-learned predictions to design competitive algorithms for online search problems.
An important property of our algorithms is that they achieve performances competitive with the offline algorithm in hindsight when the predictions are accurate (i.e., consistency) and also provide worst-case guarantees when the predictions are arbitrarily wrong (i.e., robustness). 
The proposed algorithms achieve the Pareto-optimal trade-off between consistency and robustness, where no other algorithms for online search can improve on the consistency for a given robustness. 
Further, we extend the basic online search problem to a more general inventory management setting that can capture storage-assisted energy trading in electricity markets.
In empirical evaluations using traces from real-world applications, our learning-augmented algorithms improve the average empirical performance compared to benchmark algorithms, while also providing improved worst-case performance.
\end{abstract}
\end{titlepage}

\maketitle
\section{Introduction}
With the increasing penetration of renewable energy in the supply side and distributed energy resources in the demand side, the electricity market has become increasingly volatile. 
To address the uncertainties underlying both electricity prices and energy demands in modern smart grids, competitive algorithms~\cite{borodin2005online} have been widely used for optimizing worst-case performances, such as energy scheduling in Microgrids~\cite{menati2022competitive}, storage management~\cite{mo2021optimal}, electric vehicle pricing and scheduling~\cite{bostandoost2023near}, and beyond. 

In this paper, we study competitive algorithms for energy trading problems based on a classic online search model. 
In this problem, an online decision-maker aims to sell (buy) $k\ge 1$ units of energy for the highest revenue (lowest cost) over a sequence of time-varying prices. At each step, a price is observed, and the decision-maker wants to find how many units to sell (buy) at the current price without knowing the future prices. In online decision-making, a key challenge is to balance the revenue (cost) of trading at the current price, and deferring for future better prices at the risk that those prices never arrive. 
The online search problem is foundational for modeling online decision-making in volatile markets such as asset trading in financial markets~\cite{ElYaniv2001,lorenz2009optimal,sun2021}, energy arbitrage in electricity markets~\cite{yang2020online}, and revenue management in flights market~\cite{Balseiro2023}. 

The online search problem has been tackled previously under the framework of competitive analysis~\cite{borodin2005online}. In this framework, the ultimate goal is to design online algorithms with the best possible \textit{competitive ratio}, defined as the ratio between the revenue (cost) of the offline optimum and an online algorithm. 
\citet{lorenz2009optimal} proposed two online threshold-based algorithms for both maximization and minimization versions of the online search problem. The algorithm predetermines $k$ threshold values and trades the $i$-th unit of the asset only if the price is at least (at most) equal to the $i$-th threshold value in the $k$-max ($k$-min) search. Then, by optimizing the threshold values, the online algorithms can achieve the optimal competitive ratios for both versions of $k$-search. 
The original algorithm in~\cite{lorenz2009optimal} requires that at most one unit asset can be traded in each step. Then the follow-up work~\cite{zhang2011online} extends it to a setting that allows for trading multiple units in each step with a slight modification on the algorithm by~\citet{lorenz2009optimal} using the same optimal threshold values. 
As $k$ goes to infinity, the $k$-search model in~\cite{zhang2011online} asymptotically becomes a continuous search problem, and has been framed as the one-way trading problem by the seminal work~\cite{ElYaniv2001}. 
Our paper focuses on the setting by~\cite{zhang2011online} that includes the continuous trading problem as a special case. Further, we extend the problem to an inventory management setting that is applicable to online search problems with inventory dynamics.

The classic online algorithms designed purely with guarantees of the worst-case performance tend to ignore predictions outright, and thus they often have poor performance in common average-case scenarios. In practice, however, for most application scenarios, abundant historical data could be leveraged by machine learning (ML) tools for generating some predictions of the unknown future input, e.g., prices in $k$-search.
The possibility of incorporating ML predictions in algorithmic design has led to the recent development of learning-augmented online algorithms~\cite{lykouris2018,purohit2018}, where the goal is to leverage predictions to improve the performance when predictions are accurate and preserve the robust worst-case guarantees when facing erroneous ML predictions. This high-level idea has led researchers to revisit a wide range of online problems, including but not limited to caching \cite{lykouris2018}, rent-or-buy problems \cite{purohit2018,wei2020,gollapudi2019,lee2021}, 
facility location \cite{fotakis2021,jiang2021}, 
secretary matching \cite{antoniadis2020secretary,dutting2021}, metrical task systems \cite{antoniadis2020metric}, and bin packing \cite{angelopoulos-binpacking,zeynali-online}.

Motivated by the above direction of online algorithms with predictions, this paper aims to design learning-augmented algorithms for online search problems. 
This goal is particularly crucial for applications in volatile markets where ML predictions are useful when they are accurate but can frequently advise the wrong direction, resulting in drastic losses relative to the best trade in hindsight. 

We design and analyze a learning-augmented algorithm under the consistency-robustness framework \cite{lykouris2018}, where consistency represents the competitive ratio when the prediction is accurate, and robustness is the competitive ratio regardless of the prediction error. Our goal is to design an algorithm that can achieve the Pareto-optimal trade-off between consistency and robustness, i.e., no other learning-augmented algorithms can simultaneously achieve better consistency and robustness than ours. 
Although learning-augmented algorithms have been an active research topic in recent years, the majority of prior work focuses on algorithms that can provide bounded consistency and robustness. However, studies about the Pareto-optimality of the trade-off are limited, with a few exceptions, e.g., ski-rental problem~\cite{wei2020}, online conversion problem~\cite{sun2021}, online matching problem~\cite{jin2022online} and single-leg revenue management problem~\cite{Balseiro2023}. This paper contributes to this line of research for the online search problem.

\subsection{Contributions}
\begin{figure}[t]
\centering
\includegraphics[width=.35\textwidth]{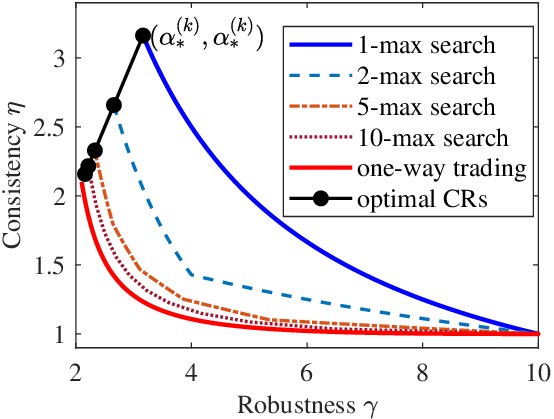} 
\includegraphics[width=.35\textwidth]{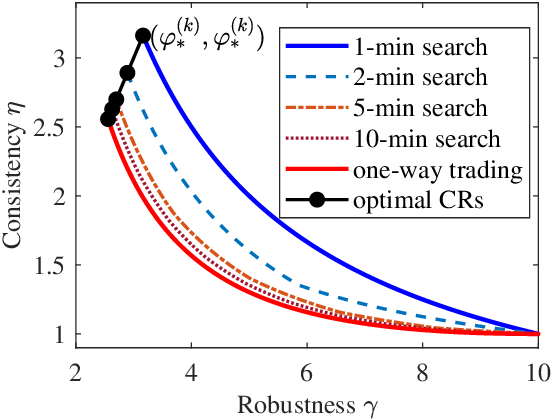}
\caption{Illustrating Pareto-optimal boundaries of $k$-max search and $k$-min search. 
All curves start from a point $(\alpha^{(k)}_*,\alpha^{(k)}_*)$ (or $(\varphi^{(k)}_*,\varphi^{(k)})_*$) with both consistency and robustness equal to the corresponding worst-case optimal competitive ratio (CR), and end at the point $(\theta,1)$. We set $\pmax = 50$, $\pmin = 5$ and $\theta = \pmax/\pmin = 10$, where $\pmax$ and $\pmin$ are upper and lower bounds of prices over the time horizon.
}
\label{fig:pareto-boundary}
\end{figure}

\paragraph{Pareto-optimal algorithms for online $k$-search problems with predictions.} We design learning-augmented algorithms for $k$-max and $k$-min search, and prove that their robustness and consistency are Pareto-optimal.
To achieve this, we start by deriving lower bound results on the trade-off between consistency and robustness of any learning-augmented algorithms. We then use the lower bound trade-off as the target of our algorithms. By leveraging ML predictions, we redesign the threshold values in the classic algorithms to prioritize the trading in cases predicted to occur for achieving good consistency while also carefully reserving sufficient trading opportunities for other cases to guarantee robustness. 
Finally, we prove the target trade-off is achievable by our design.

We note that a closely related work~\cite{sun2021} has reported Pareto-optimal algorithms for $1$-max search and one-way trading with predictions. Our proposed algorithm solves a more general $k$-search problem that includes $1$-max search (when $k=1$) and one-way trading ($k\to\infty$) as two special cases. 
In addition, we also tackle the minimization setting of $k$-search, which is demonstrated to have different performance guarantees from $k$-max search. Thus, this paper studies the full spectrum of $k$-search problems with predictions.   
In Figure~\ref{fig:pareto-boundary}, we show the Pareto boundaries achieved by our algorithms, including the results in~\cite{sun2021} as two special curves.

\paragraph{Extensions to online search with inventory dynamics.} 
In many real-world applications, the demand for trading (i.e., $k$) is unknown and is only revealed to the decision-maker over time. 
{To accommodate the online demand together with time-varying prices, we consider an online search problem assisted by an inventory; however, the inventory dynamics strongly couples the decisions over time, adding extra challenges to online decision-making. 
The presence of inventory also adds a new design space for buying and storing items in inventory when the price is low and using the stored items when the market price is not attractive. 
Despite the temporal coupling and more involved design space, we observe that the online search with inventory dynamics can be decoupled "spatially" into multiple virtual online search problems in parallel, and an independent learning-augmented algorithm can solve each of them.
Based on this observation, we extend the learning-augmented algorithms to solve this more general inventory management problem and show that the algorithm preserves the Pareto optimality when the inventory capacity is sufficiently large.
}
 
\paragraph{Applications in energy markets.} We demonstrate the performance of our learning-augmented algorithms with numerical experiments for storage-assisted energy procurement in electricity markets.
In the empirical experiments, we use energy data traces from Akamai data centers, renewable outputs from NREL, and energy prices from several ISOs.
Our proposed algorithms improve the average empirical performance compared to worst-case optimized algorithms and other baseline learning-augmented algorithms.
Moreover, our algorithms are shown to provide improved worst-case performance  
even when the predictions are with relatively large errors. Thus, our algorithms can potentially achieve best of both worlds.

\section{Problem Statement}
\subsection{Online $k$-search problem}
The $k$-max ($k$-min) search problem aims to sell (buy) $k$ units of identical items over a sequence of $T$ prices to maximize the total revenue (to minimize the total cost). 
In the online setting, the $T$ prices arrive one at a time. Upon the arrival of price $p_t$, a decision-maker must immediately decide how many items to trade, $x_t\in\{0,1,\dots,k\}$, without knowing the future prices. If the decision-maker has only traded $m$ ($m < k$) items after the arrival of the $(T-1)$-th price, she is compelled to trade all the remaining $k - m$ items at the last price $p_T$.
We make no assumptions on the underlying distribution of the prices and only assume they are bounded within known limits $\pmin$ and $\pmax$, i.e., $p_t \in [\pmin,\pmax], \forall t \in [T]$. This is a standard assumption for designing online algorithms with bounded competitive ratios in the literature of online search problems~\cite{ElYaniv2001,lorenz2009optimal}.
The price fluctuation ratio is defined as $\theta = \pmax/\pmin$ and is known to the online decision-maker in advance.

Define the price sequence $\cali := \{p_t\}_{t\in[T]}$ as an instance of the $k$-search problem. 
Let $\alg(\cali)$ denote the objective value of the $k$-search problem from an online algorithm.
Under the framework of competitive analysis~\cite{borodin2005online}, we aim to design the online algorithm such that its performance is competitive with that of an offline algorithm in hindsight.  
In particular, if an instance $\cali$ is given from the start, the $k$-max ($k$-min) search problem can be formulated as
\begin{subequations}
\label{p:k-search}
\begin{align}
    \text{max}\ (\text{min}) \quad &\sum\nolimits_{t\in[T]}p_t x_t,\\
    \text{subject\ to} \quad&\sum\nolimits_{t\in[T]} x_t = k,\\
    \text{variable}\quad & x_t \in \{0,1,\dots,k\}, \forall t\in[T].
\end{align}
\end{subequations}
Let $\opt(\cali)$ denote the optimal objective of above optimization problem.
We evaluate the performance of an online algorithm by its competitive ratio, which is defined as
\begin{align}
\label{eq:cr}
    \alpha = \max_{\cali} \frac{\opt(\cali)}{\alg(\cali)} \ \text{and}\ \varphi = \max_{\cali} \frac{\alg(\cali)}{\opt(\cali)} ,
\end{align}
for $k$-max search and $k$-min search, respectively. Both $\alpha$ and $\varphi$ are greater than one; the smaller their value, the better the performance of the online algorithm.

\subsection{Worst-case optimized algorithms} The $k$-search problem can be solved by an online threshold-based algorithm (\ota) as shown in Algorithm~\ref{alg:ota}. 
This algorithm takes $k$ threshold values $\Phi:=\{\Phi_i\}_{i\in[k]}$ in $k$-max search ($\Psi:=\{\Psi_i\}_{i\in[k]}$ in $k$-min search) as its input, and trades the $i$-th item only if the current price is at least $\Phi_i$ (at most $\Psi_i$).
Let $\ota_{\Phi}$ denote the $\ota$ with threshold $\Phi$.
$\ota_{\Phi}$ can achieve the smallest competitive ratios for $k$-search problems when the thresholds are designed optimally.
Based on the analysis in~\cite{lorenz2009optimal,zhang2011online}, we have the following lemmas. 

\begin{algorithm}[!t]
\caption{$\ota_\Phi$ for $k$-max ($\ota_\Psi$ for $k$-min) }
\label{alg:ota}
\begin{algorithmic}[1]
\State \textbf{input:} threshold values $\Phi=\{\Phi_i\}_{i\in[k]}$ ($\Psi=\{\Psi_i\}_{i\in[k]}$);
\State \textbf{initialization:} $m = 1$;
\For{step $t = 1,\dots,T-1$}
\State $x_t = 0$;
\While{$p_t \ge \Phi_m$ ($p_t \le \Psi_m$) and $m \le k$}
\State $m = m + 1$ and $x_t = x_t + 1$;
\EndWhile
\EndFor
\State $x_T = k - m + 1$.
\end{algorithmic}
\end{algorithm}

\begin{lem}[$k$-max search]
\label{lem:k-max}
$\emph{\ota}$ for $k$-max search is $\alpha^{(k)}_*$-competitive if the threshold values are
\begin{align}\label{eq:k-max}
    \Phi_i = p_{\min} \left[1 + (\alpha^{(k)}_* - 1)\left(1 + {\alpha^{(k)}_*}/{k} \right)^{i-1} \right], i\in[k],
\end{align}
where $\alpha^{(k)}_*$ is the solution of $\frac{\theta - 1}{\alpha - 1} = \left(1 + \frac{\alpha}{k}\right)^k$.
\end{lem}

\begin{lem}[$k$-min search]
\label{lem:k-min}
$\emph{\ota}$ for $k$-min search is $\varphi^{(k)}_*$-competitive if the threshold values are
\begin{align}\label{eq:k-min}
    \Psi_i = p_{\max} \left[1 - \left(1-{1}/{\varphi^{(k)}_*}\right)\left(1 + {1}/{(k\varphi^{(k)}_*)} \right)^{i-1} \right], i\in[k],
\end{align}
where $\varphi^{(k)}_*$ is the solution of $  \frac{1-1/\theta}{1-1/\varphi} = \left(1 + \frac{1}{k\varphi}\right)^k$.
\end{lem}

In the following, we sketch the key intuitions in the design and analysis of the threshold values, which are important for designing learning-augmented algorithms in the next section.
Consider the threshold values $\Phi$ for the $k$-max search. The threshold values are monotonically non-decreasing because the algorithm must aggressively trade items in the beginning to hedge the risk that values of the future prices will all drop to the lowest price $\pmin$. 
As more items are traded, the algorithm can gradually become more selective by choosing higher threshold values and taking the opportunity to trade at potentially high prices.  Since the uncertain prices vary within $[\pmin,\pmax]$, the $k$ threshold values divide this range into $k+1$ disjoint intervals $[\Phi_0, \Phi_1), [\Phi_1, \Phi_2),\dots, [\Phi_{k-1}, \Phi_{k}), [\Phi_{k}, \Phi_{k+1}]$, where $\Phi_0 := \pmin$ and $\Phi_{k+1} := \pmax$. 
Suppose the highest price falls in an interval $[\Phi_{i-1},\Phi_i)$, the total revenue of the offline algorithm is upper bounded by $k \Phi_i$, and the revenue of the online algorithm is lower bounded by $\sum_{j\in[i-1]} \Phi_j + (k-i+1)\pmin$, where the first $i-1$ items are traded at prices just equal to the first $i-1$ thresholds and the remaining items are compulsorily traded at the lowest price $\pmin$.
In this case, the ratio of the offline and online revenues is upper bounded by
\begin{align}
\label{eq:ratio-region}
\alpha_i^{(k)}(\Phi) = \frac{k \Phi_i}{\sum_{j\in[i-1]} \Phi_j + (k-i+1)\pmin}.
\end{align}
Note that Equation~\eqref{eq:ratio-region} holds for all $i\in[k+1]$ when the maximum price falls in the corresponding $k+1$ intervals.
Thus, the worst-case ratio of the $k$-max search is $\alpha^{(k)}_* = \max_{i\in[k+1]} \alpha^{(k)}_i$ and the minimum is achieved when the ratios from different intervals are balanced, i.e., ${\alpha^{(k)}_* = \alpha^{(k)}_i, \forall i\in[k+1]}$.
This gives $k+1$ equations with $k$ unknown thresholds and one unknown ratio $\alpha^{(k)}_*$. Solving those equations gives the thresholds and the competitive ratio $\alpha^{(k)}_*$ for $k$-max search in Lemma~\ref{lem:k-max}. 
We can apply similar approaches to design the threshold value $\Psi$ and the competitive ratio $\varphi^{(k)}_*$ for $k$-min search in Lemma~\ref{lem:k-min}.

\subsection{Learning-augmented algorithms}
\label{sec:learning-augmented-alg}
\paragraph{Prediction model.}
In the $k$-max ($k$-min) search, we consider a prediction $P\in[\pmin,\pmax]$ of the actual highest (lowest) price that can be obtained from ML tools. Define $\varepsilon = |V - P|$ as the error of the ML prediction, where $V$ represents the actual highest (lowest) price. 
{\color{black}
We aim to augment the worst-case optimized algorithm with the prediction $P$. 
Notably, we make no assumptions about the quality of the ML predictions, and thus the prediction error $\varepsilon$ is unknown to the algorithm. 
To evaluate the performance of the algorithm with such ``untrusted'' predictions, we focus on two metrics: (i) consistency $\eta$, which is the competitive ratio when the prediction is accurate, i.e., $\varepsilon = 0$; and (ii) robustness $\gamma$, which is the competitive ratio regardless of the prediction error. Therefore, consistency and robustness measure the algorithm's performance when the prediction is of good quality and arbitrarily bad, respectively. 
Our goal is to design the algorithm that can achieve the Pareto-optimal trade-off between consistency and robustness, i.e., for a given robustness $\gamma$, no other online algorithms can achieve a smaller consistency $\eta$.
This consistency-robustness framework was first introduced by~\cite{lykouris2018} to study the online caching problem and then has been widely adopted for designing online algorithms with untrusted predictions~\cite{purohit2018,lee2021,Balseiro2023,jin2022online}.
}

\paragraph{Baseline algorithm.} 
To design learning-augmented algorithms with bounded consistency and robustness for online resource allocation, one natural approach is to divide the limited resource into two portions, and then simultaneously run the robust algorithm and the prediction-based algorithm using the two portions, respectively (see examples in~\cite{im2021online,heinsbroek2022online}). 
Let $\lambda \in [0,1]$ be a hyper-parameter that indicates the degree of untrust in the prediction. 
The baseline algorithm divides the $k$ items into two portions, $k_r = \lceil\lambda k\rceil$ and $k_c = k - k_r$.
The algorithm, in parallel, runs a worst-case optimized $k_r$-search algorithm and a prediction-based algorithm that trades all $k_c$ items at the first price no smaller than the predicted price $P$. The decision of the baseline algorithm is the sum of the decisions from the worst-case algorithm and the prediction-based algorithm.  
\begin{lem}
\label{lem:baseline}
Given a hyper-parameter $\lambda \in [0,1]$, the baseline algorithm is $\frac{k}{k_r /\alpha^{(k_r)}_* + k-k_r}$-consistent and $\frac{k}{k_r/\alpha^{(k_r)}_* + (k-k_r)/\theta}$-robust for the learning-augmented $k$-max search problem, where $k_r:=\lceil\lambda k\rceil$.
\end{lem}

\begin{figure}[t]
\centering
\includegraphics[width=.35\textwidth]{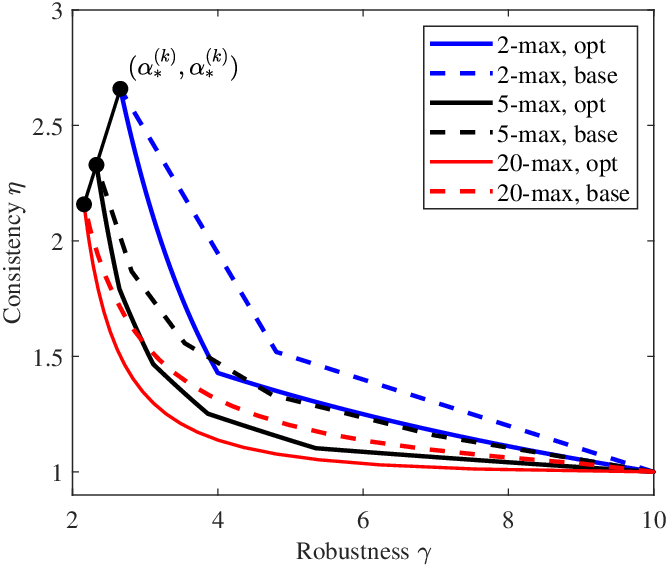}
\includegraphics[width=.35\textwidth]{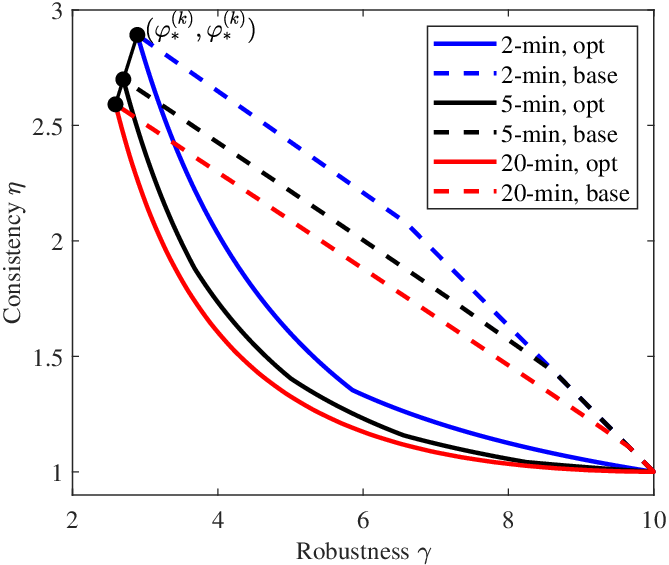} 
\caption{Comparing consistency-robustness trade-offs of baseline algorithms and Pareto-optimal algorithms for $k$-max and $k$-min search. $\pmax = 50$ and $\pmin = 5$.
}
\label{fig:comparison}
\end{figure}

Although bounded consistency and robustness can be guaranteed by the intuitive baseline algorithm, there exists a performance gap between the baseline algorithm and the Pareto-optimal algorithm as shown in Figure~\ref{fig:comparison}. 
This motivates us to design the Pareto-optimal algorithm for $k$-search to close the theoretical gap. In addition, we will show that the Pareto-optimal algorithm empirically outperforms the baseline algorithm in experiments using real data (see Section~\ref{sec:numerics}), which is of significance in real-world applications.
The proof of Lemma~\ref{lem:baseline} and the consistency-robustness result of the baseline algorithm for $k$-min search are given in Appendix~\ref{app:baseline}.

\section{Pareto-optimal Algorithms with prediction for $k$-search}

In this section, we design learning-augmented algorithms for $k$-max search and $k$-min search, and prove that they can attain the Pareto-optimal trade-off between consistency and robustness. We first focus on the $k$-max search in detail and then state the main results for $k$-min search and defer the details to Appendix~\ref{app:la}.

In $k$-max search, we have a prediction of the maximum price $P\in[\pmin,\pmax]$. 
Our learning-augmented algorithm is to redesign the threshold values $\phi(k,P) := \phi = \{\phi_i\}_{i\in[k]}$ in \ota based on the prediction $P$ such that $\ota_{\phi}$ can attain the Pareto-optimal consistency and robustness.
To achieve this goal, we first determine a lower bound for the robustness-consistency trade-off. Then, $\phi$ is designed such that $\ota_{\phi}$ can achieve this lower bound.

\subsection{Learning-augmented algorithms for $k$-max}

\subsubsection{Lower bound}
We first present a lower bound for any learning-augmented algorithm for the $k$-max problem, which is used as a target trade-off in our algorithm design.
\begin{thm}
\label{thm:lb-k-max}
For $k$-max search, any $\gamma$-robust deterministic learning-augmented algorithm  must have a consistency lower bounded by
\begin{align}
    \label{eq:lb-k-max}
    \Gamma(\gamma)=\frac{ \theta}{\left[1+(\gamma-1)(1+\frac{\gamma}{k})^{\xi}\right]/\gamma + (\theta - 1)(1-\frac{\xi}{k})},
\end{align}
where $\xi = \left\lceil{\ln\left(\frac{\theta-1}{\gamma - 1}\right)}/{\ln\left(1+\frac{\gamma}{k}\right)} \right\rceil$.
\end{thm}
The lower bound result of the $k$-max search in Theorem~\ref{thm:lb-k-max} generalizes the existing results of $1$-max search and one-way trading in~\cite{sun2021}.
When $k\to\infty$, we have $\xi/k \to \frac{1}{\gamma} \ln \frac{\theta-1}{\gamma  - 1}$ and $(1+\frac{\gamma}{k})^{\xi} \to \frac{\theta-1}{\gamma - 1}$. Thus, the lower bound of $k$-max search approaches that of one-way trading.
When $k=1$, we have $\xi = 1$. The lower bound of $k$-max search approaches that of 1-max search.

\paragraph{Proof of Theorem~\ref{thm:lb-k-max}.}
To show the lower bound, we consider a special family of instances, and show that under the special instances, any $\gamma$-robust deterministic online algorithm at least has a consistency $\eta$ that is lower bounded by $\Gamma(\gamma)$.

\begin{dfn}[$p$-instance]
A $p$-instance $\cali_p$ consists of a sequence of prices that increase continuously from $p_{\min}$ to $p$ and drop to $p_{\min}$ at the last price.  
\end{dfn}

Let $g(p):[\pmin,\pmax]\to \{0,1,\dots,k\}$ denote the cumulative trading decision of an online algorithm when it executes the instance $\cali_p$ before the compulsory trading in the last step. Since the online decision is irrevocable, $g(p)$ is non-decreasing as $p$ increases from $\pmin$ to $\pmax$. In addition, items must be traded if the price is the highest one $\pmax$. Thus, we must have $g(\pmax) = k$.
Given an online algorithm, let $\cali_{\hat{p}_i}$ denote the first instance, in which the algorithm trades the $i$-th item, i.e., $\hat{p}_i = \inf_{\{p\in[\pmin,\pmax]:g(p)\ge i\}} p$.   

For any $\gamma$-robust online algorithm, $\{\hat{p}_i\}_{i\in[k]}$ must satisfy
\begin{subequations}
\label{eq:proof-k-max1}
\begin{align}
    \frac{k \hat{p}_i}{\sum_{j\in[i-1]}\hat{p}_j +  (k-i+1) \pmin} &\le \gamma, i\in[k+1], \\
     \hat{p}_i &\le \pmax,  i\in[k],
\end{align}
\end{subequations}
where $\hat{p}_0:= \pmin$ and $\hat{p}_{k+1}:= \pmax$. 
Based on Equation~\eqref{eq:proof-k-max1}, we have 
\begin{align}
\label{eq:lem-lb-proof2}
  \hat{p}_i \le \min\left\{\pmin + \pmin(\gamma - 1)\left(1+\frac{\gamma}{k}\right)^{i-1}, \pmax\right\}, \forall i\in [k].
\end{align}

Suppose the prediction is given as $P=\pmax$. To ensure $\eta$-consistency under $\cali_{\pmax}$, any $\gamma$-robust algorithm must have 
\begin{subequations}
\begin{align}
\eta \ge \frac{\opt(\cali_{\pmax})}{\alg(\cali_{\pmax})}  = \frac{k \pmax}{\sum_{i\in[k]} \hat{p}_i} &= \frac{k \pmax}{\sum_{i\in[\xi]} \hat{p}_i + (k-\xi)\pmax}\\ 
\label{eq:lem-lb-proof}
&\ge \frac{k \pmax}{\sum_{i\in[\xi]} \pmin[1 +(\gamma  - 1)(1+\frac{\gamma}{k})^{i-1}] +  (k-\xi)\pmax}\\
&= \frac{ \theta}{\left[1+(\gamma-1)(1+\frac{\gamma}{k})^{\xi}\right]/\gamma + (\theta - 1)(1-\frac{\xi}{k})},
\end{align}   
\end{subequations}
where the inequality~\eqref{eq:lem-lb-proof} holds due to Equation~\eqref{eq:lem-lb-proof2} enforced by $\gamma$-robustness, and
$\xi$ satisfies $\pmin + \pmin(\gamma  - 1)(1+\frac{\gamma}{k})^{\xi-1} < \pmax \le  \pmin + \pmin(\gamma  - 1)(1+\frac{\gamma}{k})^{\xi}$, which gives $\xi:= \left\lceil{\ln\left(\frac{\theta-1}{\gamma - 1}\right)}/{\ln\left(1+\frac{\gamma}{k}\right)} \right\rceil$.
This completes the lower bound proof.

\subsubsection{Pareto-optimal algorithm}
Based on the lower bound result, for a given $\lambda \in [0,1]$, we set our target consistency and robustness of $k$-max search with predictions as
\begin{align}\label{eq:robust-consiste-otp}
\gamma^{(k)}(\lambda) = \alpha^{(k)}_* + (1-\lambda)(\theta - \alpha^{(k)}_*),\ \eta^{(k)}(\lambda) = \Gamma(\gamma^{(k)}(\lambda)).
\end{align}
where $\lambda$ is the confidence factor that indicates the degree of untrust in the prediction.
Our goal is to design threshold value $\phi(k,P):= \{\phi_i\}_{i\in[k]}$ that depends on $k$ and prediction $P$ such that $\ota_{\phi}$ can achieve the target in Equation~\eqref{eq:robust-consiste-otp}.
In particular, the threshold value $\phi$ is designed in the form of 
\begin{align}\label{eq:form}
\phi_i= \begin{cases}
z_i & i = 1,\dots,j^*,\\
c_i & i = j^*+1,\dots,i^*,\\
r_i & i = i^*+ 1,\dots, k,
\end{cases}
\end{align}
where the two sequences $\{z_i\}_{i\in[j^*]}$ and $\{r_i\}_{i=i^*+1}^{k}$ are designed to guarantee the robustness and the sequence $\{c_i\}_{i=j^*+1}^{i^*}$ is designed to ensure the consistency.
Recall the threshold values divide the uncertainty price range $[\pmin,\pmax]$ into $k+1$ regions and the worst-case ratio when the actual maximum price falls in each region is $\alpha_i^{(k)}(\phi) $ derived in Equation~\eqref{eq:ratio-region}. We design $\phi$ such that   
\begin{align}
\label{eq:target}
\alpha_i^{(k)}(\phi) 
\le
\begin{cases}
\eta^{(k)}, &i = j^*+1,\dots, i^*,\\
\gamma^{(k)}, &\text{otherwise},
\end{cases}
\end{align}
when prediction $P\in[\phi_{j^*+1}, \phi_{i^*}]$, and thus an accurate prediction will lead to a consistency $\eta^{(k)}$.

\paragraph{Design of threshold values.} Given a prediction $P$ of the maximum price and the target consistency $\eta$ and robustness $\gamma$, we design a piecewise threshold $\phi$ including three cases as follows. 

Let $\sigma^* \in [k]$ be the largest index such that $\frac{1 + (\theta - 1)/(1 + \gamma/k)^{k-\sigma^*}}{1 + (\eta - 1)(1+\eta/k)^{\sigma^*}} \le \frac{\gamma}{\eta}$.
We define two boundary values $\tilde{p}_1 = \pmin + \pmin(\eta - 1)(1 + \eta/k)^{\sigma^* - 1}$ and $\tilde{p}_2 = \max\{\tilde{p}_1,\gamma \pmin\}$.

\paragraph{Case I} When $P\in [\pmin, \tilde{p}_1]$, set $j^* = 0$, $i^* = \sigma^*$ and 
\begin{subequations}
\label{eq:phi-I}
\begin{align}
    c_i &= \pmin + \pmin(\eta  - 1) \left(1 + {\eta}/{k}\right)^{i-1}, i\in[i^*],\\
     r_i &= \pmin + \frac{\pmax-\pmin}{(1+\gamma/k)^{k-i+1}}, i = i^*+1,\dots,k.
\end{align}    
\end{subequations}

\paragraph{Case II} When $P\in (\tilde{p}_1,\tilde{p}_2]$, set $j^* = 0$ and 
\begin{subequations}
\label{eq:phi-II}
\begin{align}
 c_i&= \begin{cases}
    P & i = 1,\dots,m^*,\\
    \eta \frac{m^* P + (k-m^*)\pmin}{k} &  i= m^* + 1,\\
    \pmin + (c_{m^* +1} - \pmin) (1+{\eta}/{k})^{i-m^*-1} &i= m^*+2,\dots,i^*,
    \end{cases}\\
r_i &= \pmin + \frac{\pmax-\pmin}{(1+\gamma/k)^{k-i+1}}, \quad i = i^*+1,\dots,k,
\end{align}   
\end{subequations}
where $m^* = \left\lceil k\frac{P/\eta - \pmin}{P-\pmin}\right\rceil$ and $i^*$ is the largest index that ensures $\alpha_{i^*+1}^{(k)}(\phi) = \frac{k r_{i^*+1}}{\sum_{i=[i^*]} c_i + (k-i^*)\pmin}\le \gamma$.

\paragraph{Case III} When $P\in (\tilde{p}_2,\pmax]$, set
\begin{subequations}
\label{eq:phi-III}
\begin{align}
    z_i &= \pmin + \pmin(\gamma  - 1) (1 + \gamma/k)^{i-1}, \quad i\in[j^*],\\
    c_i&= \begin{cases}
    P, & i = j^*+1,\dots,m^*,\\
    \eta \frac{\sum_{i\in[j^*]} z_i + (m^* - j^*)P + (k-m^*)\pmin}{k}, & i = m^* + 1,\\ 
    \pmin + (c_{m^* +1} - \pmin) (1+{\eta}/{k})^{i-m^*-1}, & i= m^*+2,\dots,i^*,
    \end{cases}\\
    r_i &= \pmin + \frac{\pmax-\pmin}{(1+\gamma/k)^{k-i+1}}, \quad i = i^*+1,\dots,k,
\end{align}   
\end{subequations}
where we have
\begin{align*}
j^* &= \left\lceil{\ln\left(\frac{P/\pmin-1}{\gamma - 1}\right)}/{\ln\left(1+\frac{\gamma}{k}\right)} \right\rceil,\\
m^* &= j^* + \left\lceil \frac{kP/\eta - k\pmin [1 + (\gamma-1)(1+\gamma/k)^{j^*}]/\gamma}{P-\pmin}\right\rceil,
\end{align*}
and $i^*$ is the largest index such that $\alpha_{i^*+1}^{(k)}(\phi) \le \gamma$.

\begin{figure}[t]
    \subfigure[Case I with $P = 8$]{\label{fig:reservation-price}\includegraphics[width=.23\textwidth]{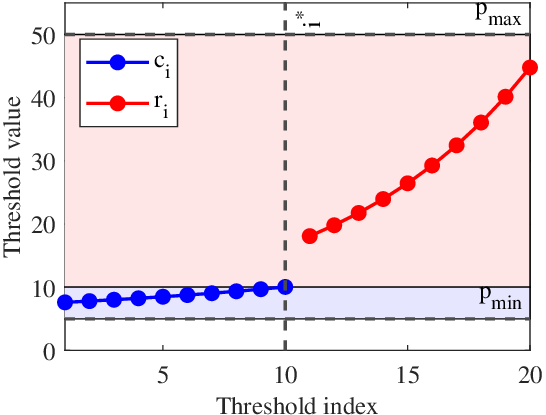}\vspace{0mm}}
    \subfigure[Case II with $P = 12$]{\label{fig:threshold-function-otp1}\includegraphics[width=.23\textwidth]{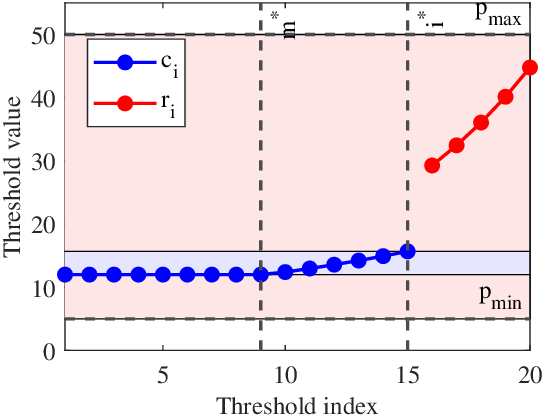}\vspace{0mm}}
    \subfigure[Case III with $P = 15$]{\label{fig:threshold-function-otp2}\includegraphics[width=.23\textwidth]{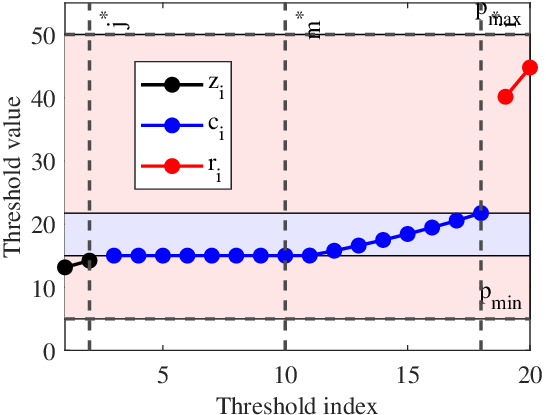}\vspace{0mm}}
    \subfigure[Case III with $P = 25$]{\label{fig:threshold-function-otp3}\includegraphics[width=.23\textwidth]{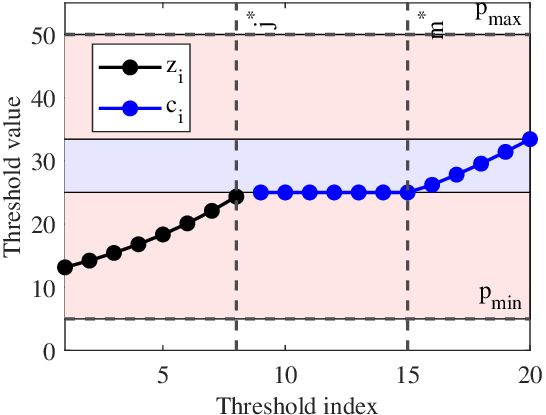}\vspace{0mm}}
    \caption{Threshold values for $k$-max search with $\pmax = 50$, $\pmin = 5$, $k = 20$, 
    $\eta = 1.52$, and $\gamma = 2.63$.}
    \label{fig:threshold}
\end{figure}
In Figure~\ref{fig:threshold}, we illustrate the design of the threshold $\phi$ in above three cases with varying predictions. $\ota_{\phi}$ can ensure that the worst-case ratio is upper bounded by $\eta$ if the actual maximum price falls in the shaded blue region and upper bounded by $\gamma$ if the actual maximum price belongs to the shaded pink region. Also note that the shaded blue region always contains the prediction $P$. Therefore, $\ota_{\phi}$ is $\eta$-consistent and $\gamma$-robust. 

\begin{thm}
\label{thm:k-max}
Given $\lambda \in[0,1]$, $\emph{\ota}_{\phi}$ is $\eta^{(k)}(\lambda)$-consistent and $\gamma^{(k)}(\lambda)$-robust for the learning-augmented $k$-max search when $\phi$ is given by Equations~\eqref{eq:phi-I}-\eqref{eq:phi-III}, where $\eta = \eta^{(k)}(\lambda)$ and $\gamma = \gamma^{(k)}(\lambda)$ are given by Equation~\eqref{eq:robust-consiste-otp}. Further, $\emph{\ota}_{\phi}$ is Pareto-optimal.
\end{thm}

We first compare the trade-off between consistency and robustness for $k$-max search with similar results in~\cite{sun2021} for $1$-max and one-way trading problems in Figure~\ref{fig:pareto-boundary}. 
In particular, we illustrate the Pareto boundaries of the consistency and robustness in one-way trading and the $k$-max search problem for four different values of $k$. 
For all problems, there exists a strong consistency-robustness trade-off. As the consistency improves from $\alpha^{(k)}_*$ (i.e., the optimal worst-case competitive ratio) to the best possible ratio $1$, the robustness degrades from $\alpha^{(k)}_*$ to the worst possible ratio $\theta$. 
In addition, as the total number of budget $k$ increases, the Pareto boundary of $k$-max search improves towards that of the one-way trading. 
This is because a large value of $k$ gives more flexibility in online decision-making to hedge the worst-case risk while using ML predictions.

We next sketch the key ideas in the design and analysis of $\ota_{\phi}$ and defer the full proof of Theorem~\ref{thm:k-max} to Appendix~\ref{app:k-max}.
Recall that in the worst-case optimized algorithm, the threshold divides the uncertainty range $[\pmin,\pmax]$ to $k+1$ intervals and the worst-case ratio $\alpha^{(k)}_i, i\in[k+1]$ is captured by Equation~\eqref{eq:ratio-region}. The algorithm adopts a balancing rule by letting $\alpha^{(k)}_i = \alpha^{(k)}_*, \forall i\in[k+1]$ to achieve the optimal competitive ratio. 
In the learning-augmented algorithm, we aim to leverage the prediction $P$ to achieve unbalanced ratios as shown in Equation~\eqref{eq:target}. 

Our high-level idea is to prioritize the trading decisions for the intervals neighbouring the prediction $P$ by designing the thresholds (i.e., $\{c_i\}_{i=j^*+1}^{i^*}$) to attain good consistency, and just guarantee robustness in above-prediction (i.e., $\{r_i\}_{i=i^*+1}^{k}$) and below-prediction (i.e., $\{z_i\}_{i\in[j^*]}$) intervals.
In particular, we design the thresholds $\{c_i\}_{i=j^*+1}^{i^*}$ by letting $\alpha^{(k)}_i \le \eta$ to ensure consistency.
The key challenge is to simultaneously ensure the robustness of below-prediction and above-prediction intervals (that are illustrated by shaded pink areas in Figure~\ref{fig:threshold}).
In the design, we have two key observations that are summarized in Propositions~\ref{pro:end} and~\ref{pro:beg}.
\begin{pro}
\label{pro:end}
Given $j \in [k-1]$, if the thresholds after $j$ are 
\begin{align}
\label{eq:pro-end}
 \phi_i = p_{\min} + \frac{p_{\max}-p_{\min}}{(1+\gamma/k)^{k-i+1}}, i = j+1,\dots, k, 
\end{align}
and $\alpha^{(k)}_{j+1}(\phi) \le \gamma$, then $\alpha^{(k)}_{i}(\phi) \le \gamma, i = j+2,\dots, k+1$.
\end{pro}
\begin{pro}
\label{pro:beg}
For a given $j \in [\xi]$, if the first $j$ thresholds are
\begin{align}
\label{eq:pro-beg}
 \phi_i = p_{\min} + p_{\min}(\gamma  - 1) \left(1 + {\gamma}/{k}\right)^{i-1}, i =1,\dots, j, 
\end{align}
then $\alpha^{(k)}_{i}(\phi) \le \gamma, i = 1,\dots, j$.
\end{pro}
First, conditioned on $\alpha^{(k)}_{j+1} \le \gamma$, if we design the remaining thresholds for $i = j+1,\dots,k$ in Equation~\eqref{eq:pro-end}, the worst-case ratios of the above-prediction intervals are upper bounded by the robustness $\gamma$, and this leads to the design of the sequence $\{r_i\}_{i=i^*+1}^{k}$ and the determination of $i^*$.  
Second, based on Proposition~\ref{pro:beg}, we can design the threshold $\{z_i\}_{i\in[j^*]}$ in Equation~\eqref{eq:pro-beg} to ensure the robustness of below-prediction intervals. 
Since the robustness of the below-prediction intervals can be trivially satisfied when prediction $P < \gamma \pmin$, the sequence $\{z_i\}_{i\in[j^*]}$ is only needed in Case III.

\subsection{Learning-augmented algorithms for $k$-min}
\label{sec:la-k-min}

In the $k$-min search, we consider a prediction ${P\in[\pmin,\pmax]}$ of the minimum price of an instance. Our algorithm is still an \ota and we aim to leverage $P$ to design the threshold {$\psi(k,P) :=\{\psi_i\}_{i\in[k]}$} such that $\ota_{\psi}$ can achieve the Pareto-optimal consistency-robustness trade-off.
We start by deriving the lower bound of the trade-off.
\begin{thm}
\label{thm:lb-k-min}
For $k$-min search, any $\gamma$-robust deterministic learning-augmented algorithm  must have a consistency lower bounded by
\begin{align}
    \Lambda(\gamma)= \theta\gamma - \theta(\gamma - 1)\left(1+{1}/{(\gamma k)}\right)^{\zeta} - (\theta - 1)\left(1 - {\zeta}/{k}\right),
\end{align}
where $\zeta = \left\lceil{\ln\left(\frac{\theta-1}{\theta - \theta/\gamma}\right)}/{\ln\left(1+ \frac{1}{\gamma k}\right)} \right\rceil$.
\end{thm}

Based on this lower bound, for a given $\lambda \in [0,1]$, we set the target consistency and robustness as 
\begin{align}
\label{eq:consist-robust-kmin}
\gamma^{(k)}(\lambda) = \varphi^{(k)}_* + (1-\lambda)(\theta - \varphi^{(k)}_*), \quad \eta^{(k)}(\lambda) = \Lambda(\gamma(\lambda)),
\end{align}
where $\varphi^{(k)}_*$ is the optimal competitive ratio of $k$-min search.

We next derive the threshold $\psi$ based on the prediction $P$.
Different from the threshold $\phi$ in the $k$-max search, $\psi$ is a non-increasing sequence of values bounded within $[\pmin, \pmax]$. $\psi$ segments the price range into $k+1$ intervals $[\psi_0, \psi_1), \dots, [\psi_{k-1}, \psi_{k}), [\psi_{k}, \psi_{k+1}]$, where $\psi_0 := \pmax$ and $\psi_{k+1} := \pmin$.
We design $\psi$ in the same form as $\phi$ in Equation~\eqref{eq:form}, where we reuse the notations in the threshold of 
$k$-max search.
Similar to $k$-max search, the threshold $\psi$ is designed such that   
\begin{align}
\label{eq:target2}
\varphi_i^{(k)}(\psi)
\le
\begin{cases}
\eta^{(k)}, &i = j^*+1,\dots, i^*,\\
\gamma^{(k)}, &\text{otherwise},
\end{cases}
\end{align}
where $\varphi_i^{(k)}(\psi) := \frac{\sum_{j\in[i-1]} \psi_j + (k-i+1)\pmax}{k \psi_i}$ when the prediction of the minimum price $P\in[\psi_{j^*+1}, \psi_{i^*}]$. Thus, an accurate prediction will lead to consistency $\eta$, and regardless of the prediction, the robustness $\gamma$ can be guaranteed. $\psi$ is also designed in three cases, and the Pareto-optimality of $\ota_{\psi}$ is given in the follow-up theorem.  

Let $\sigma^* \in [k]$ be the largest index that $\frac{1 - (1 - 1/\eta)(1+{1}/{(\eta k)})^{\sigma^*}}{1 - (1 - 1/\theta)/(1 + {1}/{(\gamma k)})^{k - \sigma^*}} \le \frac{\gamma}{\eta}$.
We define two boundary values $\tilde{p}_1 = \pmax - \pmax(1 - 1/\eta)(1 + 1/(\eta k))^{\sigma^* - 1}$ and $\tilde{p}_2 = \min\{\tilde{p}_1, \pmax/\gamma\}$.

\paragraph{Case IV} When $P\in (\tilde{p}_1, \pmax]$, set $j^* = 0$, $i^* = \sigma^*$, and 
\begin{subequations}
\label{eq:psi-I}
\begin{align}
c_i &= \pmax - \pmax\left(1  - {1}/{\eta}\right) \left(1 +{1}/{(\eta k)}\right)^{i-1}, \quad i\in[i^*],\\
 r_i &= \pmax - \frac{\pmax-\pmin}{(1+ 1/(\gamma k))^{k-i+1}}, \quad i = i^*+1,\dots,k.
\end{align}    
\end{subequations}

\paragraph{Case V} When $P\in (\tilde{p}_2, \tilde{p}_1]$, set $j^* = 0$ and 
\begin{subequations}
\label{eq:psi-II}
\begin{align}
c_i&= \begin{cases}
P, & i = 1,\dots,m^*,\\
 \frac{m^* P + (k-m^*)\pmax}{\eta k}, &  i= m^* + 1,\\
\pmax - (\pmax - c_{m^* +1}) \left(1 + {1}/{(\eta k)} \right)^{i-m^*-1}, & i= m^*+2,\dots,i^*,
\end{cases}\\
r_i &= \pmax - \frac{\pmax-\pmin}{(1+ 1/(\gamma k))^{k-i+1}}, \quad i = i^*+1,\dots,k,
\end{align}    
\end{subequations}
where $m^* = \left\lceil k\frac{\pmax- \eta P }{\pmax-P}\right\rceil$ and $i^*$ is the largest index that ensures
$\varphi^{(k)}_{i^*+1}(\psi)\le \gamma.$

\paragraph{Case VI} When $P\in [\pmin, \tilde{p}_2]$, set
\begin{subequations}
\label{eq:psi-III}
\begin{align}
z_i & = \pmax - \pmax\left(1 - {1}/{\gamma}  \right) \left(1+{1}/{(\gamma k)}\right)^{i-1},\quad i\in[j^*],\\
c_i&= \begin{cases}
P, & i = j^*+1,\dots,m^*,\\
\frac{\sum_{i\in[j^*]} z_i + (m^* - j^*)P + (k-m^*)\pmax}{\eta k },& i = m^* + 1,\\ 
\pmax - (\pmax - c_{m^* +1}) \left(1 + {1}/{(\eta k)}\right)^{i-1}, & i= m^*+2,\dots,i^*,
\end{cases}\\
r_i &= \pmax - \frac{\pmax-\pmin}{(1+ 1/(\gamma k))^{k-i+1}},\quad i = i^*+1,\dots,k,
\end{align}    
\end{subequations}
in which we have $j^* = \left\lceil{\ln\left(\frac{1 - P/\pmax}{1 - 1/\gamma }\right)}/{\ln\left(1+{1}/{(\gamma k)}\right)} \right\rceil$, and\\ $m^* = j^* + \lceil \frac{k\pmax  [1 - (1 - 1/\gamma)(1+1/(\gamma k))^{j^*}] - kP\eta}{\pmax - P}\rceil,$ and $i^*$ is the largest index such that $\varphi^{(k)}_{i^*+1}(\psi) \le \gamma$.

\begin{thm}
\label{thm:k-min}
Given $\lambda \in[0,1]$, $\emph{\ota}_{\psi}$ is $\eta^{(k)}(\lambda)$-consistent and $\gamma^{(k)}(\lambda)$-robust for the learning-augmented $k$-min search when $\psi$ is given by Equations~\eqref{eq:psi-I}-\eqref{eq:psi-III}, where $\eta := \eta^{(k)}(\lambda)$ and $\gamma :=\gamma^{(k)}(\lambda)$ are given in Equation~\eqref{eq:consist-robust-kmin}. Further, $\emph{\ota}_{\psi}$ is Pareto-optimal.
\end{thm}

\section{Online Search with Inventory Dynamics}
\label{sec:osid}
In this section, we consider an online search problem with inventory dynamics, extend the learning-augmented algorithm for online search to this more general problem, and prove the Pareto-optimality of this algorithm. 

\subsection{Problem statement}
\label{sec:osid-p}

Consider an inventory management problem in a time-slotted system with horizon length $T$.
In a volatile market with time-varying prices, an inventory with capacity $B$ is used to store items when the price is low and use inventory to satisfy demand when the price is high. 
In each slot $t$, we first observe a demand $d_t \in \{0,1,\dots\}$ that must be fulfilled immediately and a price $p_t$ for purchasing the items in the market.  
The demand $d_t$ can be satisfied by dual sources, either by the inventory that was purchased in previous slots, or the items purchased from the market.
Then we decide $x_t \in \{0,1,\dots\}$, the number of items to purchase from the market, in each slot $t$. If $x_t > d_t$, the additional items $x_t - d_t$ are stored in the inventory for future use.
The goal is to minimize the total cost of purchasing items while fulfilling all real-time demands. 

Let $\cali:=\{p_t, d_t\}_{t\in[T]}$ denote an instance of the online search with inventory dynamics (\osid). Given $\cali$, the problem can be formulated as the following integer linear program.
\begin{subequations}
\label{p:osid}
\begin{align}
    \label{eq:osid-obj}
    \text{(\texttt{min-OSID})}\quad\min \quad &\sum\nolimits_{t\in[T]}p_t x_t,\\
    \label{eq:osid1}
    \text{subject\ to} \quad&s_t = s_{t-1} + x_t - d_t, \quad\forall t\in[T],\\
    \label{eq:osid2}
    & x_t \ge d_t - s_{t-1}, \quad\forall t\in[T],\\
    \label{eq:osid3}
    & s_t \le B, \quad\forall t\in [T],\\
    \label{eq:osid4}
    \text{variables}\quad & x_t, s_t \in \{0,1,\dots\}, \quad\forall t\in[T].
\end{align}
\end{subequations}
where $s_t$ is the inventory level at the end of slot $t$ with initial state $s_0 = 0$. The objective~\eqref{eq:osid-obj} of \osid is to minimize the total cost of purchasing the items over $T$ slots. 
The constraint~\eqref{eq:osid1} enforces the inventory dynamics, i.e., the inventory level $s_t$ is equal to the previous inventory level $s_{t-1}$ plus the net amount of purchased items $x_t - d_t$. The constraints~\eqref{eq:osid2} and~\eqref{eq:osid3} ensure that the inventory level is within capacity, i.e., $s_t \ge 0$ and $s_t \le B$ for $t\in[T]$.

{\osid is a general framework that can capture various resource allocation problems with inventory dynamics. For instance, we can apply \osid to model and solve the storage-assisted energy trading problem in this paper.
Particularly, consider a large energy consumer (e.g., data centers) that aims to purchase energy from the real-time energy market. 
It is equipped with an energy storage that can be used to store energy when the energy price is low and discharge to satisfy energy demand when the price is high.
In each slot $t$, the consumer observes its energy demand $d_t$ and the real-time electricity price $p_t$, and then decides how much energy to purchase, $x_t$, such that the demand can be satisfied by the combined amount from purchasing and storage discharging. 
In energy trading, demand $d_t$ and decision $x_t$ may be relaxed to continuous variables. Our algorithms and results can be extended for the continuous version of \osid (See Appendix~\ref{app:osid}).
}

From the theoretical perspective, the offline $\osid$ problem~\eqref{p:osid} is an extension of the $k$-min search formulated in~\eqref{p:k-search}.
Let $\bar{k} = \min_{t\in[T]: d_t >0} d_t$ denote the minimum non-zero demand, and without loss of generality, consider $B \ge \bar{k}$ for an inventory management problem.
When $d_t = 0$ for $t = 1,\dots, T-1$, and $d_T = \bar{k}$, \osid is to minimize the cost of buying a total of $\bar{k}$ items and storing them in the inventory over the first $T-1$ slots, and to use the inventory to satisfy the demand $\bar{k}$ in the last slot. This reduces to a $\bar{k}$-min search problem. When facing the general time-varying demand, \osid is much challenging than the classical online search problem. 

{
\color{black}
Correspondingly, the $k$-max version of \osid can be used to model the problem of selling renewable energy (from solar panels or wind turbines) in electricity markets. 
The seller is equipped with an energy storage and can use it to temporally store the renewable energy when the market price is not attractive, waiting for a better selling opportunity. 
In each slot $t$, the seller obtains $a_t\in\{0,1,\dots\}$ amount of renewable energy with no production costs, and observes the real-time electricity price $p_t$. It then determines to sell $x_t$ amount of energy in the market and earns a profit $x_t p_t$. If $x_t > a_t$, the additional energy $x_t - a_t$ is discharged from the energy storage. Otherwise, the unsold energy $a_t - x_t$ is stored in the inventory for trading in the future. The overall problem can be cast as follows:     
\begin{subequations}
\label{p:osid-max}
\begin{align}
    \label{eq:osid-max-obj}
    \text{(\texttt{max-OSID})}\quad\max\quad &\sum\nolimits_{t\in[T]}p_t x_t,\\
    \label{eq:osid-max1}
    \text{subject\ to} \quad&s_t = s_{t-1} - x_t + a_t, \quad\forall t\in[T],\\
    \label{eq:osid-max2}
    & x_t \le a_t + s_{t-1}, \quad\forall t\in[T],\\
    \label{eq:osid-max3}
    & s_t \le B, \quad\forall t\in [T],\\
    \label{eq:osid-max4}
    \text{variables}\quad & x_t, s_t \in \{0,1,\dots\}, \quad\forall t\in[T].
\end{align}
\end{subequations}
Note that if the renewable energy has a marginal production cost $c_t$ ($c_t < \pmin, \forall t\in[T]$), we can use a refined electricity price $\hat{p}_t := p_t - c_t$ with refined price uncertainty range $[\hat{p}_{\min}, \hat{p}_{\max}]$, where $\hat{p}_{\min} := \pmin - \max_{t\in[T]} c_t$ and $\hat{p}_{\max} := \pmax - \min_{t\in[T]} c_t$ in the \texttt{max-OSID}. For simplicity of presentation, we focus on \texttt{min-OSID} in the rest of the paper; but the algorithms and results can be straightforwardly extended for \texttt{max-OSID}.
}

\subsection{Learning-augmented algorithm for \osid}

\subsubsection{\osid with predictions.}
The key idea for designing online algorithms for \osid is to note the following fact: in each slot $t$, if the inventory level is sufficiently large to cover the demand $d_t$, we can use the inventory to satisfy the demand first, and then run a virtual online search algorithm to buy $d_t$ items from the volatile market in the following slots and restore them back to the inventory. 
Thus, by running multiple virtual search problems in parallel, an online algorithm for \osid can achieve a similar performance to that of the online search if the inventory is non-empty. 

Based on the above idea, we categorize time slots into busy periods with $s_t > 0$ and idle periods with $s_t = 0$, and make purchasing decisions differently in busy and idle periods. In particular, given an online algorithm for \osid, we divide the time horizon into a total of $N$ intervals, where each interval $n$ starts at a slot with non-empty inventory and ends before the starting of the next interval or $T$. Each interval $n$ can be further divided into a busy period followed by an idle period. Let $\calt_{n}^+$ and $\calt_{n}^0$ denote sets of slots in the busy 
period and idle period of the interval $n$, respectively. 

In the beginning of each slot $t$, we are given a prediction $P_t$. When $t\in \calt_{n}^+$, $P_t$ is the predicted minimum price in the busy period of interval $n$. When $t\in \calt_{n}^0$, $P_t$ is a prediction of the minimum price in the idle period of interval $n$ and the busy period of $n+1$.
Overall the prediction $P_t$ is the minimum price that a virtual search problem created at slot $t$ can observe in its lifecycle.

\subsubsection{Learning-augmented algorithm} We propose an online algorithm for \osid in Algorithm~\ref{alg:osid}. 
This algorithm maintains multiple virtual search problems in parallel in the busy period of each interval and uses the algorithm for $k$-min search with predictions in Section~\ref{sec:la-k-min} as a subroutine to solve each virtual search problem.
In the following, we call the $k$-min search with prediction $P$ as a $(k, P)$-search problem.
Let $\psi(k,P) := \{\psi_i(k,P)\}_{i\in[k]}$ denote the threshold values for $(k, P)$-search that can attain Pareto-optimal consistency-robustness trade-off in Equation~\eqref{eq:consist-robust-kmin}. In the algorithm, we set a target robustness guarantee $\gamma$, and choose the optimal consistency $\eta^{(k)} := \eta^{(k)}(\gamma)$ for given $\gamma$ and $k$.

Algorithm~\ref{alg:osid} divides the time horizon into $N$ intervals.
For each $t\in\calt_{n}^+$ in the busy period, the algorithm observes the demand $d_t$ and price $p_t$, and receives the prediction $P_t$. It then creates a virtual $(d_t, P_t)$-search problem if demand is strictly positive $d_t > 0$.
Then the algorithm runs the learning-augmented online threshold-based algorithm (\ota) for all the $i$ created online search problems in parallel and obtains $\{x_t^{(n,j)}\}_{j\in[i]}$, where $x_t^{(n,j)}$ is the online decision of the $j$-th search problem. 
The online decision of \osid is then set to the maximum of $\sum_{j\in[i]} x_{t}^{(n,j)}$ (i.e., the sum of trading decisions of all virtual search problem) and $d_t - s_{t-1}$ (i.e., the minimum trading amount to satisfy the demand). If the first term is larger, we have $s_t > 0$ and the busy period continues. Otherwise, the inventory becomes empty and the algorithm enters the idle period. 
Then the algorithm clears all virtual search problems created in the busy period, and initiates a $(B, P_t)$-search problem based on the physical capacity of the inventory, waiting for a low price that can start the busy period of the next interval.

\begin{algorithm}[t]
\caption{Learning-augmented algorithm for \osid}
\label{alg:osid}
\begin{algorithmic}[1]
\State \textbf{input:} threshold $\psi(k,P)$, inventory $B$, target robustness $\gamma$;
\State\textbf{initiation:} $n = 1$, $i = 1$; $m^{(1,1)} = 1$, $k^{(1,1)} = B$, $P^{(1,1)} = P_0$, $s_0 = 0$; $\psi^{(1,1)} := \psi(k^{(1,1)},P^{(1,1)})$;
\label{alg:adj-line-init-storage}
\For{$t = 1, \dots, T$}
\State observe demand $d_t$ and price $p_t$; obtain prediction $P_t$;
\If{$d_t > 0$} 
\State $i = i+1$, $k^{(n,i)} = d_t$, $P^{(n,i)} = P_t$, $\psi^{(n,i)} := \psi(k^{(n,i)},P^{(n,i)}) = \{\psi^{(n,i)}_j\}_{j\in[k^{(n,i)}]}$, $m^{(n,i)} = 1$;
\EndIf
\For{$j = 1,\dots, i$} 
\label{alg:virtual-start}
\State $x_{t}^{(n,j)} = 0$;
\While{$p_t \le \psi^{(n,j)}_{m^{(n,j)}}$ and $m^{(n,j)} \le k^{(n,j)}$}
\label{alg:adj-line-subroutine-call}
\State $m^{(n,j)} = m^{(n,j)} + 1$ and $x_{t}^{(n,j)} = x_{t}^{(n,j)} + 1$;
\EndWhile
\EndFor
\label{alg:virtual-end}
\State $x_t = \max\left\{\sum_{j\in[i]} x_{t}^{(n,j)}, d_t - s_{t-1}\right\}$; \label{alg-line:online-sol} 
\State $s_t = s_{t-1} + x_t - d_t$;
\label{alg-line:dynamics}
\If{$s_t = 0$ and $s_{t-1} > 0$} 
\State $n = n + 1$, $i = 1$, $m^{(n,1)} = 1$, $k^{(n,1)} = B$, $P^{(n,1)} = P_t$, $\psi^{(n,1)} := \psi(k^{(n,1)},P^{(n,1)})$;
\EndIf
\EndFor
\end{algorithmic}
\end{algorithm}

\subsubsection{Consistency-robustness analysis}
We next show that Algorithm~\ref{alg:osid} provides the best consistency-robustness guarantees. 
We still assume all prices are bounded $p_t\in[\pmin,\pmax]$, and in \osid, we additionally make the following assumption for the demand. 
\begin{ass}\label{ass:demand}
    The average demand in the idle period of one interval is smaller than capacity $B$, i.e., $\frac{1}{N}\sum_{n\in[N]}\sum_{t\in\calt^0_n} d_t \le B$. 
\end{ass}
This assumption basically means \osid is equipped with a sufficiently large inventory such that on average we can buy and store cheap items during busy periods and use the inventory to satisfy the demand in idle periods with high prices.

\begin{thm}
\label{thm:osid}
Under Assumption~\ref{ass:demand}, given $\lambda \in[0,1]$, Algorithm~\ref{alg:osid} is $\eta^{(\bar{k})}(\lambda)$-consistent and $\gamma^{(\bar{k})}(\lambda)$-robust for the learning-augmented \osid when $\psi(k,P)$ is given by Equations~\eqref{eq:psi-I}-\eqref{eq:psi-III}, where $\eta^{(\bar{k})}(\lambda)$ and $\gamma^{(\bar{k})}(\lambda)$ are defined in Equation~\eqref{eq:consist-robust-kmin}, and $\bar{k} = \min_{t\in[T]: d_t >0} d_t$ is the minimum non-zero demand.
Algorithm~\ref{alg:osid} is Pareto-optimal for \osid with minimum non-zero demand $\bar{k}$.
\end{thm}
Our main result in this section is to show that our proposed learning-augmented algorithms for \osid can achieve the best possible consistency-robustness trade-off, and this trade-off is the same as that of $\bar{k}$-min search with predictions, where $\bar{k}$ is the minimum non-zero demand of \osid and unknown to algorithm a prior.
The key step of Algorithm~\ref{alg:osid} is to, in parallel, run multiple learning-augmented \ota for online search problems, and thus can be considered as 
an extension of the \ota for online search with predictions.
The consistency-robustness guarantee is restricted by the performance of the online search problem with capacity $\bar{k}$ since $\bar{k}$-min search is a special case of \osid with minimum non-zero demand $\bar{k}$.  



\subsubsection{Proof of Theorem~\ref{thm:osid}}
We sketch the proof for Theorem~\ref{thm:osid}, and defer all details to Appendix~\ref{app:inventory}.
Let $\alg(\cali)$ and $\opt(\cali)$ denote the costs from the online algorithm and the offline optimal algorithm under an instance $\cali$.
We first show that Algorithm~\ref{alg:osid} generates a feasible solution for \osid.

\begin{lem}\label{lem:feasible}
    The online solution of Algorithm~\ref{alg:osid} is feasible for \osid.
\end{lem}


Let $i_n$ denote the number of virtual search problems created in the busy period of interval $n$, and let $m^{(n,i)} -1$ denote the total number of items purchased by the $i$-th virtual search problem in interval $n$.
The next lemma shows that we can upper bound the total cost of the algorithm by threshold values $\{\psi^{(n,i)}_j\}_{n\in[N],i\in[i_n],j\in[m^{(n,i)}-1]}$ from the virtual search problems.

\begin{lem}\label{lem:ub-alg-oi}
The cost of Algorithm~\ref{alg:osid} for \osid is upper bounded by
\begin{align*}
\emph{\alg}&(\cali) \le \left[s_T + R\right] p_{\max} \\
&+ \sum_{n\in[N]}\sum_{i\in[i_n]} \biggl[ \sum_{j\in [m^{(n,i)}-1]} \psi^{(n,i)}_j 
+ (k^{(n,i)} - m^{(n,i)} + 1) p_{\max}
\biggr],
\end{align*}
where $R = \sum_{n\in[N]}\sum_{t\in\calt_n^0} d_t - NB$.
\end{lem}

Next we show that the total cost of the offline algorithm is lower bounded by $\{\psi^{(n,i)}_{m^{(n,i)}}\}_{n\in[N],i\in[i_n]}$ from the virtual search problems.

\begin{lem}\label{lem:lb-opt-oi}
The cost of offline optimum for \osid is lower bounded
\begin{align}
\emph{\opt}(\cali) \ge \sum_{n\in[N]}\sum_{i\in[i_n]} 
\psi^{(n,i)}_{m^{(n,i)}}\cdot k^{(n,i)} + \frac{R \cdot p_{\max}}{\gamma}. 
\end{align}
\end{lem}

Based on the property of threshold values given in~\eqref{eq:target2}, regardless of the accuracy of the predictions, we have $\sum_{j\in [m^{(n,i)}-1]} \psi^{(n,i)}_j + (k^{(n,i)} - m^{(n,i)} + 1) p_{\max} \le \gamma \psi^{(n,i)}_{m^{(n,i)}} k^{(n,i)}, \forall n\in[N], i\in[i_n]$.
Combining Lemma~\ref{lem:ub-alg-oi} and Lemma~\ref{lem:lb-opt-oi} gives
 \begin{align*}
\frac{\alg(\cali) - s_T p_{\max}}{\opt(\cali)} 
&\le 
\frac{\gamma\sum\nolimits_{n\in[N]}\sum\nolimits_{i\in[i_n]} 
\psi^{(n,i)}_{m^{(n,i)}} k^{(n,i)} + R p_{\max} }{\sum\nolimits_{n\in[N]}\sum\nolimits_{i\in[i_n]} 
\psi^{(n,i)}_{m^{(n,i)}} k^{(n,i)} + \frac{R \cdot p_{\max}}{\gamma}} = \gamma.
\end{align*}   
Thus, the robustness of Algorithm~\ref{alg:osid} is given by $\alg(\cali) \le \gamma \opt(\cali) + s_T \pmax$, where $s_T \pmax$ is an additive loss due to the remaining inventory at the end of time horizon. Note that this additive loss is inevitable compared to the offline optimum since the offline algorithm knows the length of time horizon, and thus can always use up all inventory at the end of horizon to minimize the cost. 
Further, as the time horizon $T$ increases, both $\alg(\cali)$ and $\opt(\cali)$ increase and the performance loss is dominated by the multiplier term $\gamma$ while the additive term $s_T \pmax$ is negligible.

When all predictions are accurate, the threshold values in Equations~\eqref{eq:psi-I}-\eqref{eq:psi-III} ensure that
$$\sum_{j\in [m^{(n,i)}-1]} \psi^{(n,i)}_j + (k^{(n,i)} - m^{(n,i)} + 1) p_{\max} \le \eta^{(k^{(n,i)})} \psi^{(n,i)}_{m^{(n,i)}} k^{(n,i)}, \forall n\in[N], i\in[i_n].$$
Based on Lemma~\ref{lem:ub-alg-oi} and Lemma~\ref{lem:lb-opt-oi}, we have 
\begin{align*}
\frac{\alg(\cali) - s_T p_{\max}}{\opt(\cali)} 
\le \frac{\eta^{(\bar{k})}\sum\nolimits_{n\in[N]}\sum\nolimits_{i\in[i_n]}  
\psi^{(n,i)}_{m^{(n,i)}} k^{(n,i)} + R p_{\max} }{\sum\nolimits_{n\in[N]}\sum\nolimits_{i\in[i_n]} 
\psi^{(n,i)}_{m^{(n,i)}} k^{(n,i)} + \frac{R \cdot p_{\max}}{\gamma}}\le \eta^{(\bar{k})},
\end{align*}
where the last inequality holds since $R =\sum_{n\in[N]}\sum_{t\in\calt_n^0} d_t - NB \le 0$ under the 
Assumption~\ref{ass:demand}.
Thus, Algorithm~\ref{alg:osid} can guarantee $\alg(\cali) \le \eta^{(\bar{k})} \opt(\cali) + s_T \pmax$.

Therefore, Algorithm~\ref{alg:osid} is $\gamma^{(\bar{k})}$-robust and $\eta^{(\bar{k})}$-consistent. Also, note that $\bar{k}$-min search with predictions is a special case of \osid with minimum non-zero demand $\bar{k}$. $\gamma^{(\bar{k})}$ and $\eta^{(\bar{k})}$ are also the lower bound of the consistency-robustness trade-off. Thus, the algorithm is Pareto-optimal.

\subsection{Inventory-cost-aware algorithms}
\label{sec:extension}
{
In numerous real-world applications of \osid, the utilization of inventory introduces additional expenses, such as the battery cost and energy conversion loss in the context of the storage-assisted energy trading problem. To address these associated costs, we extend Algorithm~\ref{alg:osid} to an inventory-cost-aware algorithm. 
Specifically, our extension accounts for two types of inventory-related costs in the energy trading problem: (i) \textit{usage cost}, entailing a cost of $\delta$ for each unit of energy charged into or discharged from inventory; and (ii) \textit{conversion loss}, wherein only $\rho_c$ and $\rho_d$ percentages of energy can be efficiently charged into and discharged from the inventory, respectively.     
The inventory-cost-aware problem can be cast as:
\begin{subequations}
   \begin{align}
\label{eq:obj}
\min_{x_t \ge 0, s_t \ge 0} \quad& \sum\nolimits_{t\in[T]} p_t x_t + |d_t - x_t|\delta\\
\label{eq:dynamics}
    \text{subject\ to}\quad& s_t = s_{t-1} + \rho_c x^c_t -  \frac{1}{\rho_d} x^d_t, \forall t\in[T]\\
    &x^c_t = (x_t - d_t)^+, \forall t\in[T],\\
    &x^d_t = (d_t - x_t)^+, \forall t\in[T],\\
    & d_t - x_t \le \rho_d s_{t-1}, \forall t\in[T],\\
    & s_t \le B, \forall t\in [T],
\end{align} 
\end{subequations}
where the objective~\eqref{eq:obj} additionally considers a cost $|d_t - x_t|\delta$ from inventory usage and the inventory dynamics~\eqref{eq:dynamics} takes into account the energy conversion loss. We further relax the integral constraints for this energy trading problem.

We make two modifications in Algorithm~\ref{alg:osid} to handle inventory costs. 
To deal with the conversion loss, we increase the capacity of the \textit{first} virtual storage to $k^{(1,1)} = {B}/{(\rho_c\rho_d)}$ in Line~\ref{alg:adj-line-init-storage}. The intuition for adjusting only the first storage is that its extended capacity of $B$ can be considered largely responsible for moving units in and out of the physical storage.  Modifying just this storage will adjust the charging amount and leaving the virtual storages with capacity $d_t$ untouched will maintain trading to meet demand.

To deal with the inventory usage cost, we modify the threshold prices depending on the current demand. 
The primary objective is to encourage the algorithm to refrain from utilizing the inventory unless it has good potential to reduce the purchasing cost substantially.
To achieve this, we set thresholds to avoid purchasing a quantity too close to $d_t$. Specifically, we modify the thresholds $\psi^{(n,j)}$ in Line~\ref{alg:adj-line-subroutine-call} to use $\psi'^{(n,j)}$ defined as follows:
\begin{align*}
\psi_i'^{(n,j)}  &= \begin{cases}
    \psi_{\tau}^{(n,j)}-2\delta, & \psi_{\tau}^{(n,j)} - 2\delta \leq \psi_i^{(n,j)} < \psi_{\tau}^{(n,j)}\\
    \psi_{\tau}^{(n,j)}, & \psi_{\tau}^{(n,j)} \leq \psi_i^{(n,j)} \leq \psi_{\tau}^{(n,j)} +2\delta\\
    \psi_i^{(n,j)}, & \text{otherwise};\\
    \end{cases} 
\end{align*}
where $\tau = m^{(n,j)} + d_t \cdot \left\lfloor\frac{k^{(n,j)}}{
\sum_{v}k^{(n,v)}}\right\rfloor$.
First, we set a target utilization $\tau$ such that the aggregate target utilization over $j$ inventories is $d_t$ more than the current level.  Then, prices within a range $2\delta$ of the corresponding target price $\psi_{\tau}^{(n,j)}$ are modified such that the algorithm only charges to the physical storage if the price is at least $2\delta$ above/below the threshold price for meeting demand $d_t$.  The range $2\delta$ is chosen because it inevitably costs $2\delta$ to charge and discharge one unit.
Adding usage costs and conversion loss fundamentally alters the problem formulation of \osid, making it much more challenging to analyze the theoretical performance of the inventory-cost-aware algorithm. Therefore, we leave the theoretical analysis of the algorithm for our future work but verify the performance numerically in Section~\ref{sec:numerics}.
}

\section{Experimental Results}
\label{sec:numerics}
\subsection{Experimental setup}
We apply the learning-augmented algorithms for the continuous \osid (see Appendix~\ref{app:osid} for more details). For supplementary experiments on the basic $k$-search problem, see Appendix~\ref{sec:bitcoin}.
We set the length of each slot to 5 minutes and set the time horizon of each instance to 24 hours, i.e., $T = 288$.  The energy storage capacity is set to $18 \times \max_{t \in [T]}d_t$ such that a full charge of the storage can power the data center for 90 minutes of maximum energy demand. We report the empirical ratios over 31 days.

\noindent\textbf{Energy price.} We use the day-ahead electricity prices of local electricity markets over several different independent system operators in the United States, i.e., CAISO, NYISO, ERCOT, and ISO-NE \cite{iso-data}.  To have a consistent 5-minute settlement interval, we up-sample some market price readings.

\noindent\textbf{Data center demand.} We use a repository of demand traces from Akamai's server clusters collected during a 31-day period from multiple locations.  We run our algorithms on data centers located in relevant electricity markets: New York for NYISO, Dallas for ERCOT, Los Angeles for CAISO, and Boston for ISO-NE. The energy demand is partially supplied by renewable generation, which has uncertainty that causes the net demand (the data center demand minus renewable supply) to be more volatile.

\noindent\textbf{Algorithm predictions.} The learning-augmented algorithm for \osid utilizes a prediction $P_t$ that predicts the minimum price in the lifecyle of the virtual online search created at slot $t$. In our experiments, we consider utilizing the minimum price of half-hour, 1-hour, and 2-hour windows. To adjust the prediction quality, we add prediction errors drawn from a normal distribution with mean 0 and variance $\sigma$ scaled by a factor of $\theta$.

\begin{table*}[!t]
	\caption{The reported numbers are empirical cost ratios (the lower, the better) of algorithms in varying markets and seasons; $\theta$ is the fluctuation ratio; $\alpha^{(\infty)}_*$ is the theoretical competitive ratio.}
	\label{tbl:comp}
	\centering
	\scriptsize
	\setlength\tabcolsep{2pt}
	\begin{tabular}{|c|c|c|c|c|c|c||c|c|c||c|c|c||c|c|c|}
		\hline
		\multirow{2}{*}{}&\multirow{2}{*}{\textbf{Season}} & \multirow{2}{*}{$\theta$} & \multirow{2}{*}{$\alpha^{(\infty)}_*$} & \multirow{2}{*}{\nostr} & \multirow{2}{*}{\arp} & \multicolumn{3}{|>{\centering}m{2.1cm}|}{\textbf{Half-hour window}} & \multicolumn{3}{|>{\centering}m{2.1cm}|}{\textbf{1-hour window}} & \multicolumn{3}{|>{\centering}m{2.1cm}|}{\textbf{2-hour window}}\\
		\hhline{~~~~~~----------}
		&&&&&  & \osbase & \osalf &  \osstatic   & \osbase & \osalf &  \osstatic   & \osbase & \osalf &  \osstatic  \\
		\hline \hline
		\multirow{5}{*}{\rotatebox[origin=c]{90}{ CAISO}} & \textbf{Spring} & 33.25 & 4.40 & 1.34 & 1.29 & 1.31 & 1.19 & 1.11 & 1.26 & 1.22 & 1.17 & 1.24 & 1.24 & 1.22   \\
		
		&\textbf{Summer} & 44.59 & 5.05 & 1.47 & 1.37 & 1.36 & 1.21 & 1.12 & 1.31 & 1.23 & 1.15 & 1.40 & 1.41 & 1.38   \\
		
		&\textbf{Fall} & 4.62 & 1.83 & 1.36 & 1.23 & 1.26 & 1.18 & 1.10 & 1.22 & 1.15 & 1.10 & 1.22 & 1.17 & 1.16  \\
		
		&\textbf{Winter} & 18.78 & 3.39 & 1.43 & 1.26 & 1.27 & 1.18 & 1.09 & 1.26 & 1.17 & 1.14 & 1.26 & 1.22 & 1.20  \\
		\hline\hline
		&\textbf{Year} & \textbf{25.31} & \textbf{3.67} & \textbf{1.40} & \textbf{1.28} & \textbf{1.30} & \textbf{1.19} & \textbf{1.11} & \textbf{1.26} & \textbf{1.19} & \textbf{1.14} & \textbf{1.28} & \textbf{1.25} & \textbf{1.24} \\
		
		\hline\hline
		\multirow{5}{*}{\rotatebox[origin=c]{90}{ NYISO}} & \textbf{Spring} & 3.74 & 1.68 & 1.36 & 1.20 & 1.22 & 1.15 & 1.11 & 1.19 & 1.16 & 1.13 & 1.18 & 1.20 & 1.19  \\
		
		&\textbf{Summer} & 5.93 & 2.03 & 1.36 & 1.28 & 1.28 & 1.22 & 1.12 & 1.23 & 1.17 & 1.11 & 1.23 & 1.22 & 1.20   \\
		
		&\textbf{Fall} & 4.66 & 1.84 & 1.38 & 1.28 & 1.28 & 1.21 & 1.13 & 1.23 & 1.18 & 1.13 & 1.21 & 1.20 & 1.19   \\
		
		&\textbf{Winter} & 2.83 & 1.50 & 1.21 & 1.17 & 1.19 & 1.13 & 1.07 & 1.21 & 1.14 & 1.11 & 1.20 & 1.17 & 1.14 \\
		\hline\hline
		&\textbf{Year}& \textbf{4.29} & \textbf{1.76} & \textbf{1.31} & \textbf{1.22} & \textbf{1.23} & \textbf{1.17} & \textbf{1.10} & \textbf{1.21} & \textbf{1.16} & \textbf{1.12} & \textbf{1.20} & \textbf{1.20} & \textbf{1.17}  \\	
		
		\hline \hline
		\multirow{5}{*}{\rotatebox[origin=c]{90}{ ERCOT}} & \textbf{Spring} & 7.66 & 2.27 & 1.38 & 1.28 & 1.29 & 1.20 & 1.10 & 1.28 & 1.22 & 1.12 & 1.30 & 1.31 & 1.30 \\
		
		&\textbf{Summer} & 4.98 & 1.89 & 1.52 & 1.26 & 1.26 & 1.19 & 1.11 & 1.25 & 1.20 & 1.11 & 1.22 & 1.21 & 1.20   \\
		
		&\textbf{Fall} & 50.37 & 5.34 & 1.57 & 1.41 & 1.40 & 1.21 & 1.11 & 1.39 & 1.28 & 1.16 & 1.42 & 1.40 & 1.39   \\
		
		&\textbf{Winter} & 9.13 & 2.45 & 1.33 & 1.26 & 1.27 & 1.22 & 1.12 & 1.24 & 1.18 & 1.13 & 1.21 & 1.20 & 1.19 \\
		\hline\hline
		&\textbf{Year} & \textbf{18.03} & \textbf{2.99} & \textbf{1.45} & \textbf{1.30} & \textbf{1.30} & \textbf{1.20} & \textbf{1.11} & \textbf{1.29} & \textbf{1.22} & \textbf{1.13} & \textbf{1.28} & \textbf{1.28} & \textbf{1.27}  \\\hline \hline
		
		\multirow{5}{*}{\rotatebox[origin=c]{90}{ISO-NE}} & \textbf{Spring} & 10.57 & 2.62 & 1.32 & 1.26 & 1.27 & 1.25 & 1.10 & 1.25 & 1.24 & 1.13 & 1.24 & 1.23 & 1.19 \\
		
		&\textbf{Summer} & 9.37 & 2.48 & 1.47 & 1.34 & 1.33 & 1.22 & 1.11 & 1.30 & 1.24 & 1.17 & 1.34 & 1.32 & 1.28 \\
		
		&\textbf{Fall} & 6.70 & 2.15 & 1.36 & 1.27 & 1.29 & 1.18 & 1.10 & 1.27 & 1.22 & 1.16 & 1.28 & 1.27 & 1.22 \\
		
		&\textbf{Winter} & 46.44 & 5.14 & 1.69 & 1.43 & 1.39 & 1.29 & 1.13 & 1.36 & 1.34 & 1.25 & 1.32 & 1.32 & 1.31 \\
		\hline\hline
		&\textbf{Year} & \textbf{18.27} & \textbf{3.10} & \textbf{1.46} & \textbf{1.33} & \textbf{1.32} & \textbf{1.23} & \textbf{1.11} & \textbf{1.29} & \textbf{1.26} & \textbf{1.18} & \textbf{1.30} & \textbf{1.29} & \textbf{1.25}  \\\hline
	\end{tabular}
\end{table*}

\begin{figure*}[t]
\centering
\subfigure[Effect of prediction quality on settings of $\lambda$]{
\includegraphics[width=.3\textwidth]{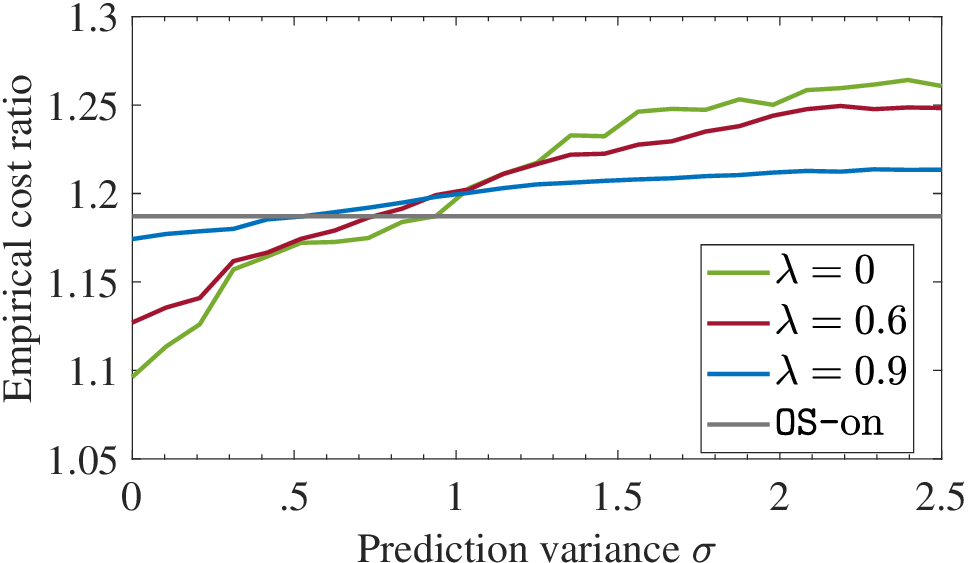} \label{fig:osid-preds}}
\subfigure[Impact of inventory usage cost]{
\includegraphics[width=.3\textwidth]{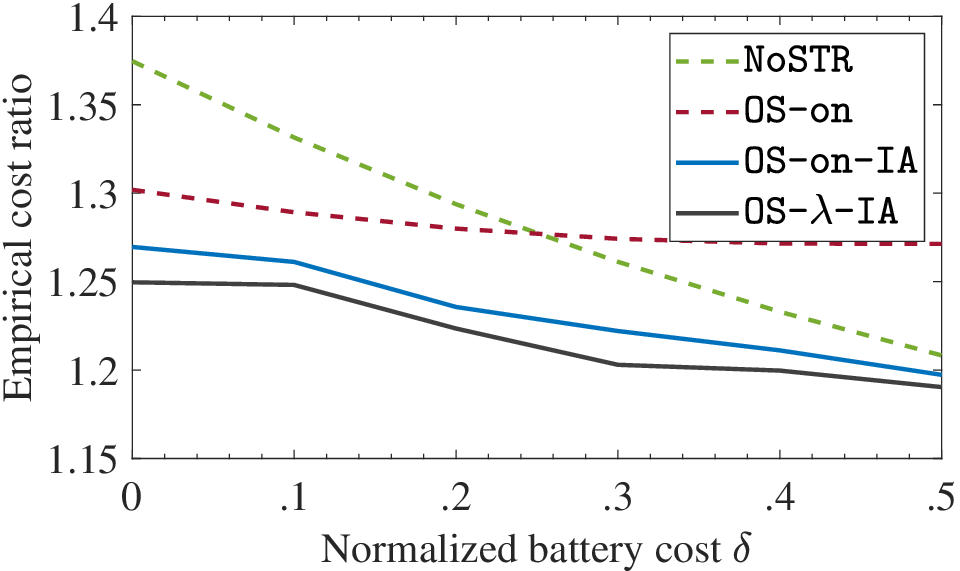} \label{fig:osid-bat}}
\subfigure[Impact of conversion loss]{
\includegraphics[width=.3\textwidth]{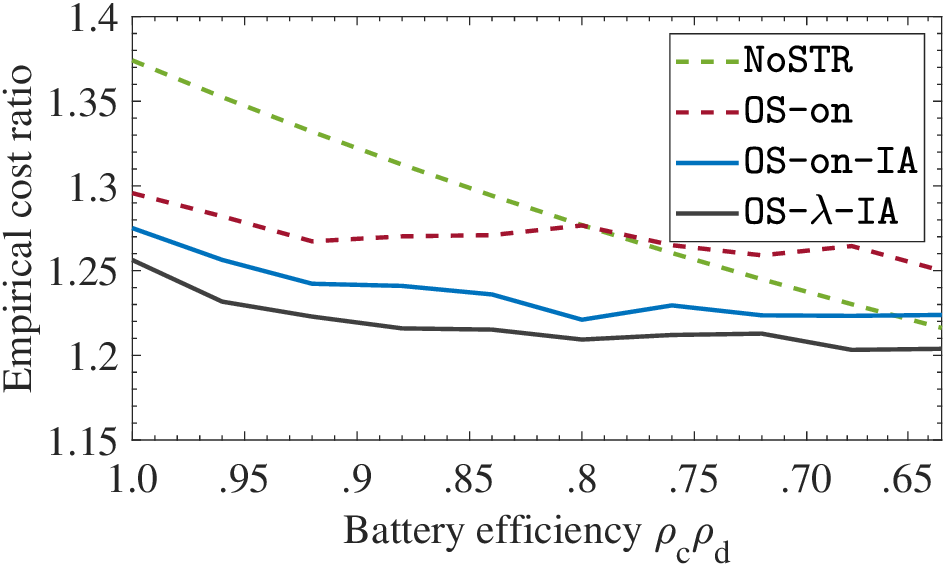} \label{fig:osid-loss}}
\caption{Cost ratio vs. different problem parameters.  Dallas Jan 2023 with half-hour prediction window.
}

\end{figure*}

\subsection{Comparison algorithms}

We compare the following seven algorithms. (\nostr): An algorithm that satisfies the demand by purchasing energy from the market without using storage. 
The cost ratio of \nostr demonstrates the maximum benefit of storage.
(\arp): The worst-case optimized online algorithm~\cite{yang2020online} that solves \osid without predictions.
(\osbase): The baseline algorithm introduced in Section~\ref{sec:learning-augmented-alg} adapted for \osid.
(\osalf): Our proposed learning-augmented algorithm for \osid.  The value of $\lambda$ is selected according to an online learning algorithm (i.e., adversarial Lipschitz algorithm~\cite{maillard2010}) to adaptively select $\la$. 
(\osstatic): Our proposed algorithm for \osid with the best choice of $\lambda$. 
\revision{
(\osalfia): Our extended learning-augmented algorithm that is aware of inventory usage costs and conversion loss.
(\arpia): The inventory-cost-aware algorithm without predictions.}

\subsection{Experimental results}

We report the average empirical ratios of five algorithms in Table \ref{tbl:comp} across data center locations, seasons, and prediction window sizes.
The magnitude of the price fluctuation ratio $\theta$ clearly indicates the problem difficulty, especially for \arp. For example, in the seasonal breakdown of ISO-NE, the cost ratio of \arp increases by $13\%$ in winter as compared to spring.  The same effect can be observed for learning-augmented algorithms to a lesser degree: the same comparison for \osalf increases by $3 \%$ from spring to winter.

We observe that \osalf outperforms other baseline algorithms averaged over the course of a year.  The degree of performance gain depends on the prediction quality, with smaller prediction windows yielding a greater gain.  In the half-hour prediction window, \osalf outperforms \arp averaged over a year by between 5\% to 8\%, depending on the dataset. In comparison, the 2-hour window yields a year-average performance gain between 1.5\% to 3\%.  
The performance of \osalf relative to \osbase highlights the importance of the optimality of our algorithm over all other prediction-based algorithms, with a performance improvement of up to 8.4\% in the year-average CAISO half-hour trial. Both \osalf and \osstatic have relatively smaller improvements on \osbase in the 2-hour window. Despite this, all prediction-based algorithms perform better than \arp, indicating that even with poor predictions, good usage of the trust parameter still improves on worst-case optimized algorithms.

{
To demonstrate the inherent tradeoff between the robustness and consistency of our learning-augmented algorithms, we show the performance of the algorithms with varying prediction errors in Figure~\ref{fig:osid-preds}.  
The prediction error is generated by Gaussian noise $\epsilon \sim \mathcal{N}(0,\sigma)$, where $\sigma \in [0,2.5]$ and is normalized by $\theta$.  
Recall that algorithms with different $\lambda$s achieve different consistency-robustness tradeoffs, a smaller $\lambda$ rendering a better consistency and worse robustness.  
The algorithm with $\lambda = 0$ attains the best consistency, i.e., it achieves the lowest cost ratio when the prediction error is small; however, its robustness is the worst, i.e., it has the highest cost ratio as the prediction error becomes large. 
As we increase the value of $\lambda$, the algorithm can experience lower cost ratios in scenarios with large prediction errors, at the cost of higher cost ratios when the prediction error is small. This corresponds with the theoretical robustness-consistency tradeoffs.
Moreover, the performances of all learning-augmented algorithms smoothly improve with increasing prediction quality. This is desirable for algorithms with untrusted predictions, and it underscores that, despite the algorithms' obliviousness to the prediction quality, their performances consistently improve as prediction accuracy increases. 
}

{
In Figure~\ref{fig:osid-bat} and Figure~\ref{fig:osid-loss}, we showcase the impacts of inventory usage cost and conversion loss on proposed algorithms. We compare the average cost ratios of our inventory-cost-aware algorithms against \nostr and \arp, which are ignorant of the additional inventory dynamics. The usage cost $\delta$ in Figure~\ref{fig:osid-bat} is normalized by $\pmin$ and the charging and discharging losses are represented by the overall battery efficiency $\rho_c\rho_d$.  Storage efficiency in datacenters depends on the energy storage device used, ranging from as high as $95\%$ for ultra-capacitors or as low as $68\%$ for compressed air energy storage ~\cite{wang2012}.
We observe that the increase in the usage cost or the conversion loss tends to lower the cost ratios of algorithms since the cost of the offline algorithm also rises. As usage costs or energy loss grow, offline algorithm charges the battery less frequently, and \nostr becomes increasingly viable.  
\osalfia offers consistent performance improvements over baseline algorithms with different battery costs until $\delta = 0.5$, where it matches the cost ratio of \nostr.  As $\delta$ increases, the performance gain from using \osalfia over \arp grows from 5\% to almost 10\%.  For battery efficiency below 65\%, utilizing predictions is key to maintaining improvement over baselines, as \arpia matches \nostr.
}

\section{Conclusion}
We have designed online algorithms augmented by machine-learned predictions for the online search problem and its extension to online search with inventory dynamics. 
Theoretically, the performance of our proposed algorithms is consistent with that of offline algorithms in hindsight when the prediction is accurate and robust on prediction errors. 
Further, the trade-off between consistency and robustness has been proven Pareto-optimal. 
Practically, we have applied the learning-augmented algorithms to the storage-assisted energy trading problem in energy markets.
Through extensive experiments using real traces, our proposed learning-augmented algorithm has been shown to achieve the best of both worlds in the sense that it improves the average empirical performance compared to existing benchmark algorithms, while also improving the worst-case performance even when the predictions are inaccurate.

\section{Acknowledgment}
The work of Russell Lee and Mohammad Hajiesmaili was supported by the U.S. National Science Foundation (NSF) under grant numbers CAREER-2045641, NGSDI-2105494, CPS-2136199, CNS-2106299, and CNS-2102963.  The work of John C.S. Lui was supported in part by the RGC GRF 14202923.

\bibliographystyle{ACM-Reference-Format}
\bibliography{references}

\newpage
\onecolumn
\appendix
\section{Proofs on baseline algorithms}
\label{app:baseline}
{

We first prove Lemma~\ref{lem:baseline} that provides the consistency and robustness of the baseline algorithm for $k$-max search. For completeness of the presentation, we then give the corresponding result for $k$-min search that can be derived similarly. 

\subsection{Proof of Lemma~\ref{lem:baseline}}
\label{app:proof-baseline-k-max}
Given a hyper-parameter $\lambda \in [0,1]$, the baseline algorithm reserves $k_r = \lceil\lambda k\rceil$ capacity and $k_c = k - k_r$ capacity for running the robust algorithm and the prediction-based algorithm, respectively.
In particular, the robust algorithm is the worst-case optimized algorithm (i.e., Algorithm~\ref{alg:ota} with threshold values given in Equation~\eqref{eq:k-max}) and the prediction-based algorithm waits to trade all $k_c$ items until the first price that is no smaller than the predicted maximum price $P$.

Let $\bar{x}_t$ and $\hat{x}_t$ denote the trading decisions from the robust algorithm and the prediction-based algorithm, respectively. Then the baseline algorithm trades $x_t = \bar{x}_t + \hat{x}_t$ in step $t$.
Given an instance $\cali = \{p_t\}_{t\in[T]}$, the total revenue of the baseline algorithm is $\alg(\cali)= \sum\nolimits_{t\in[T]} p_t x_t = \sum\nolimits_{t\in[T]} p_t \bar{x}_t + \sum\nolimits_{t\in[T]} p_t \hat{x}_t$ and the offline optimum is $\opt(\cali) = k V$, where $V = \max_{t\in[T]} p_t$ is the actual maximum price. 

We first show the robustness of the baseline algorithm. 
Since $\bar{x}_t$ is given by the optimal $k_r$-max search algorithm, we have $\sum\nolimits_{t\in[T]} p_t \bar{x}_t \ge k_r V /\alpha^{(k_r)}_*$, in which $\alpha^{(k_r)}_*$ is the optimal competitive ratio of $k_r$-max search. In addition, the revenue of the prediction-based algorithm is lower bounded by $\sum\nolimits_{t\in[T]} p_t \hat{x}_t \ge L k_c \ge k_c V /\theta$. Therefore, we have
\begin{align}
    \alg(\cali) \ge k_r V /\alpha^{(k_r)}_* + k_c V /\theta = \left[k_r /(k\alpha^{(k_r)}_*) + (k-k_r) /(k\theta) \right]\opt(\cali).
\end{align}
Thus, the robustness of the baseline algorithm is $k/[k_r/\alpha^{(k_r)}_* + (k-k_r)/\theta]$.

Next, we prove the consistency. When the prediction $P$ is accurate, i.e., $P = V$, the revenue of the prediction-based algorithm is $\sum\nolimits_{t\in[T]} p_t \hat{x}_t = V k_c$. We then have
\begin{align}
    \alg(\cali) \ge k_r V /\alpha^{(k_r)}_* + k_c V  = \left[k_r /(k\alpha^{(k_r)}_*) + (k-k_r) /k \right]\opt(\cali).
\end{align}
Thus, the consistency is $k/[k_r/\alpha^{(k_r)}_* + k - k_r]$. This completes the proof. 

\subsection{Results of baseline algorithms for $k$-min search.}
The baseline algorithm can also solve the $k$-min search problem. We just need to modify the robust algorithm to the optimal $k_r$-min search algorithm and the prediction-based algorithm to wait to trade all $k_c$ items until the first price that is no larger than the predicted minimum price $P$. Then we can have the following result.
\begin{lem}
\label{lem:baseline-min}
Given $\lambda \in [0,1]$, the baseline algorithm is $\frac{k_r\varphi^{(k_r)}_* + k-k_r}{k}$-consistent and $\frac{k_r\varphi^{(k_r)}_* + (k-k_r)\theta}{k}$-robust for the learning-augmented $k$-min search problem, where $k_r:=\lceil\lambda k\rceil$.
\end{lem}

\paragraph{Proof of Lemma~\ref{lem:baseline-min}.} Given an instance $\cali = \{p_t\}_{t\in[T]}$, the total cost of the baseline algorithm is $\alg(\cali)= \sum\nolimits_{t\in[T]} p_t x_t = \sum\nolimits_{t\in[T]} p_t \bar{x}_t + \sum\nolimits_{t\in[T]} p_t \hat{x}_t$ and the offline optimum is $\opt(\cali) = k V$, where $V = \min_{t\in[T]} p_t$ is the actual minimum price. 

We first prove the robustness. 
Since $\bar{x}_t$ is from the optimal $k_r$-min search algorithm, we have $\sum\nolimits_{t\in[T]} p_t \bar{x}_t \le k_r V \varphi^{(k_r)}_*$, in which $\varphi^{(k_r)}_*$ is the optimal competitive ratio of $k_r$-min search. In addition, the cost of the prediction-based algorithm is upper bounded by $\sum\nolimits_{t\in[T]} p_t \hat{x}_t \le \pmax k_c \le k_c V \theta$. Therefore, we have
\begin{align}
    \alg(\cali) \le k_r V\varphi^{(k_r)}_* + k_c V \theta = \opt(\cali) [k_r \varphi^{(k_r)}_*  + (k-k_r] \theta  )/k.
\end{align}
Thus, the robustness of the baseline algorithm is $[k_r\varphi^{(k_r)}_* + (k-k_r)\theta]/k$.

Next, we prove the consistency. When the prediction $P$ is accurate, i.e., $P = V$, the cost of the prediction-based algorithm is $\sum\nolimits_{t\in[T]} p_t \hat{x}_t = V k_c$. We then have
\begin{align}
    \alg(\cali) \le k_r V \varphi^{(k_r)}_* + k_c V  = \opt(\cali)[k_r \varphi^{(k_r)}_*/k + (k-k_r) /k ].
\end{align}
Thus, the consistency is $(k_r\varphi^{(k_r)}_* + k-k_r)/k$.

}

\section{Proofs on Pareto-optimal learning-augmented algorithms}
\label{app:la}
\subsection{Proof of Theorem~\ref{thm:k-max}}
\label{app:k-max}

The threshold values $\phi = \{\phi_i\}_{i\in[k]}$ for $k$-max is a non-decreasing sequence $\phi_1\le\phi_2 \le \dots\le \phi_k$. For notations convenience, we define $\phi_0 = \pmin$ and $\phi_{k+1} = \pmax$ .
The threshold $\phi$ divides the uncertainty range $[\pmin,\pmax]$ into $k+1$ regions.
If $\phi_i = \phi_{i+1} = \dots = \phi_{j} < \phi_{j+1}$, we use $j+1$ to index the interval $[\phi_i, \phi_{j+1})$.
If \ota totally trades $i$ items (excluding the compulsory trading at the last step), then the worst-case price sequence is 
\begin{align*}
    \phi_1, \phi_2, \dots, \phi_{i},\underbrace{\phi_{i+1}-\epsilon,\dots, \phi_{i+1}-\epsilon}_{k}, \underbrace{\pmin,\dots,\pmin}_k,
\end{align*}
where $\epsilon > 0$ and $\epsilon \to 0$. 
The worst-case ratio under this price sequence can be characterized by
\begin{align}
    \alpha_{i+1}^{(k)}(\phi) = \frac{k\phi_{i+1}}{\sum_{j\in[i]} \phi_{j} + (k-i) \pmin}.
\end{align}

Before proceeding to the proof of Theorem~\ref{thm:k-max}, we first prove Proposition~\ref{pro:end} and Proposition~\ref{pro:beg}.

\paragraph{Proof of Proposition~\ref{pro:end}.}
Given $j\in[k-1]$, if $\alpha^{(k)}_{j+1} \le \gamma$, we have
\begin{align}
\label{eq:pro-proof1}
    \sum\nolimits_{s\in[j]}\phi_s + (k-j)\pmin \ge \frac{k \phi_{j+1}}{\gamma}.
\end{align}
We then have
\begin{subequations}
\begin{align}
    \alpha^{(k)}_{j+2} = \frac{k \phi_{j+2}}{\sum_{s\in[j+1]}\phi_s + (k - j - 1)\pmin}  &= \frac{k \phi_{j+2}}{\sum_{s\in[j]}\phi_s + (k - j)\pmin + \phi_{j+1} - \pmin}\\
    \label{eq:pro-proof2}
    &\le \frac{k \phi_{j+2}}{k \phi_{j+1}/\gamma + \phi_{j+1} - \pmin}\\
    \label{eq:pro-proof3}
    &= \frac{\gamma \phi_{j+2}}{ \pmin + (1+{\gamma}/{k})[\phi_{j+1} - \pmin]} = \gamma,
\end{align}   
\end{subequations}
where \eqref{eq:pro-proof2} holds due to Inequality~\eqref{eq:pro-proof1} and the last equality~\eqref{eq:pro-proof3} is based on the design of the threshold values in Equation~\eqref{eq:pro-end}.
By following the same step, the worst-case ratios in the following intervals can be shown that $\alpha_i^{(k)} \le \gamma, i = j+3,\dots, k+1$. 

\paragraph{Proof of Proposition~\ref{pro:beg}.}
When the first $j$ threshold values are designed based on Equation~\eqref{eq:pro-beg}, the worst-case ratio of the $i$-th interval of \ota can be shown as
\begin{align}
    \alpha^{(k)}_i = \frac{k \phi_{i}}{ \sum_{s\in[i-1]}\phi_{s} + (k-i + 1)\pmin} = \frac{k \phi_{i}}{ \frac{k}{\gamma}[\pmin + \pmin(\gamma - 1)(1 + \gamma/k)^{i-1}]} = \gamma, \forall i\in [j].
\end{align}

In the following, we show \ota with the designed threshold $\phi$ can guarantee the target consistency and robustness in all three cases.

\paragraph{Case I} The prediction is $P\in[\pmin,\tilde{p}_1]$, where $\tilde{p}_1 = \pmin + \pmin(\eta - 1)(1 + \eta/k)^{\sigma^* - 1}$ and $\sigma^* \in [k]$ is the largest index such that 
\begin{align}
\label{eq:cond1}
    \frac{1 + (\theta - 1)/(1 + \gamma/k)^{k-\sigma^*}}{1 + (\eta - 1)(1+\eta/k)^{\sigma^*}} \le \frac{\gamma}{\eta}. 
\end{align}
We first prove that there exists an index $\sigma^* \in [k]$ when the target consistency and robustness are set based on Equation~\eqref{eq:robust-consiste-otp}. 
It is sufficient to show that Equation~\eqref{eq:cond1} holds when $\sigma^* = 1$, i.e., 
\begin{align}\label{eq:proof1}
    \frac{1 + (\theta - 1)/(1 + \gamma/k)^{k-1}}{1 + (\eta - 1)(1+\eta/k)} \le \frac{\gamma}{\eta}.
\end{align}

When $1+(\gamma - 1)(1 + \gamma/k)^{k-1} \ge \theta$, Equation~\eqref{eq:proof1} holds since we can have
\begin{align*}
    \frac{1 + (\theta - 1)/(1 + \gamma/k)^{k-1}}{1 + (\eta - 1)(1+\eta/k)} \le \frac{1 + (\theta - 1)/(1 + \gamma/k)^{k-1}}{\eta} \le \frac{\gamma}{\eta}.
\end{align*}

When $1+(\gamma - 1)(1 + \gamma/k)^{k-1} < \theta$, we have $\xi = \left\lceil{\ln\left(\frac{\theta-1}{\gamma - 1}\right)}/{\ln\left(1+\frac{\gamma}{k}\right)} \right\rceil = k$ and the corresponding target consistency is $\eta = \frac{\theta \gamma}{1 + (\gamma - 1)(1+\gamma/k)^k}$. Then we can construct a function $f(\gamma) = \frac{\gamma}{\eta} [1 + (\eta - 1)(1+\eta/k)] - [1 + (\theta - 1)/(1 + \gamma/k)^{k-1}]$. By noting that $f(\gamma)$ is increasing in $[\alpha^{(k)}_*,\infty)$ and $f(\alpha^{(k)}_*) = 0$, we have $f(\alpha^{(k)}_*) \ge 0$, which gives Equation~\eqref{eq:proof1}. 

In the following, we start to prove the consistency and robustness by showing that $\alpha^{(k)}_i(\phi)$ satisfies Equation~\eqref{eq:target}. 
We can derive the worst-case ratios $\alpha^{(k)}_i(\phi)$ given the design of $\phi$ as 
\begin{subequations}
\begin{align*}
    c_i &= \pmin + \pmin(\eta  - 1) \left(1 + {\eta}/{k}\right)^{i-1}, i\in[i^*],\\
     r_i &= \pmin + \frac{\pmax-\pmin}{(1+\gamma/k)^{k-i+1}}, i = i^*+1,\dots,k.
\end{align*}    
\end{subequations}  
For $i\in[i^*]$, we have
\begin{align*}
    \alpha^{(k)}_i = \frac{k c_i}{\sum_{j\in[i-1]} c_{j} + (k-i+1) \pmin} = \frac{k c_i}{ k c_i /\eta }= \eta.
\end{align*}

For $i = i^* + 1$, we have
\begin{align*}
    \alpha^{(k)}_{i^*+1} = \frac{k r_{i^*+1}}{\sum_{j\in[i^*]} c_{j} + (k-i^*) \pmin} &= \frac{k \left[\pmin + {(\pmax-\pmin)}/{(1+\gamma/k)^{k-i^*}}\right]}{\frac{k}{\eta} \left[\pmin + \pmin(\eta  - 1) \left(1 + {\eta}/{k}\right)^{i^*} \right] }\\
    &= \frac{ \eta \left[1 + {(\theta-1)}/{(1+\gamma/k)^{k-\sigma^*}}\right]}{1 + (\eta  - 1) \left(1 + {\eta}/{k}\right)^{\sigma^*}} \le \gamma,
\end{align*}
where the last inequality is due to Equation~\eqref{eq:cond1}.

For $i = i^*+2,\dots, k+1$, based on Proposition~\ref{pro:end} and $\alpha^{(k)}_{i^*+1} \le \gamma$, we have $\alpha^{(k)}_i \le \gamma,$ for all $i = i^*+2,\dots, k+1$.

Summarizing above results, if $P\in[\pmin,\tilde{p}_1]$, we have $\alpha^{(k)}_i \le \eta, i\in[i^*]$ and thus $\ota_{\phi}$ is $\eta$-consistent. The worst-case competitive ratio is $\max_{i\in[k+1]} \alpha^{(k)}_i =\gamma$, and thus $\ota_{\phi}$ is $\gamma$-robust.

\paragraph{Case II} When the prediction $P\in (\tilde{p}_1,\tilde{p}_2]$, where $\tilde{p}_2 = \max\{\tilde{p}_1, \gamma \pmin\}$. This case exists only when $\tilde{p}_1 < \gamma \pmin$.
In the following, we derive the worst-case ratios $\alpha^{(k)}_i$ given the threshold design 
\begin{subequations}
\begin{align*}
 c_i&= \begin{cases}
    P & i = 1,\dots,m^*,\\
    \eta \frac{m^* P + (k-m^*)\pmin}{k} &  i= m^* + 1,\\
    \pmin + (c_{m^* +1} - \pmin) (1+{\eta}/{k})^{i-m^*-1} &i= m^*+2,\dots,i^*,
    \end{cases}\\
r_i &= \pmin + \frac{\pmax-\pmin}{(1+\gamma/k)^{k-i+1}}, \quad i = i^*+1,\dots,k,
\end{align*}   
\end{subequations}
where $m^* = \left\lceil k\frac{P/\eta - \pmin}{P-\pmin}\right\rceil$ and $i^*$ is the largest index that ensures $\alpha_{i^*+1}^{(k)}(\phi) = \frac{k r_{i^*+1}}{\sum_{i=[i^*]} c_i + (k-i^*)\pmin}\le \gamma$.

For $i = 1$, $\alpha^{(k)}_1 = k P/ (k \pmin)  \le \tilde{p}_2/\pmin \le \gamma$. 

For $i = m^*+1$ (noting that $c_i = P, \forall i\in[m^*]$), 
\begin{align*}
    \alpha^{(k)}_{m^*+1} = \frac{k c_{m^*+1}}{m^* P + (k-m^*)\pmin} = \eta.
\end{align*}

For $i= m^* + 2, \dots, i^*$, we have
\begin{align*}
    \alpha^{(k)}_{i} &= \frac{k c_{i}}{m^* P + \sum_{j=m^*+1}^{i-1} c_j + (k-i+1)\pmin}= \frac{k c_{i}}{\frac{k}{\eta} [\pmin + (c_{m^*+1}-\pmin)(1+\eta/k)^{i - m^*-1}]} = \eta.
\end{align*}

For $i = i^* + 1, \dots, k+1$, based on $\alpha^{(k)}_{i^*+1} \le \gamma$ and Proposition~\ref{pro:end}, we have $\alpha^{(k)}_i \le \gamma$.

Based on above results, when prediction $P\in (\tilde{p}_1,\tilde{p}_2]$, we have $\alpha^{(k)}_i \le \eta, i= m^* + 1,\dots, i^*$, and thus the consistency of $\ota_{\phi}$ is $\eta$. Since $\max_{i\in[k+1]} \alpha^{(k)}_i \le \gamma$, the robustness is $\gamma$.

\paragraph{Case III} When the prediction $P\in (\tilde{p}_2,\pmax]$, the worst-case ratios $\alpha^{(k)}_i$ can be shown as follows when the threshold values are given by
\begin{subequations}
\begin{align*}
    z_i &= \pmin + \pmin(\gamma  - 1) (1 + \gamma/k)^{i-1}, \quad i\in[j^*],\\
    c_i&= \begin{cases}
    P, & i = j^*+1,\dots,m^*,\\
    \eta \frac{\sum_{i\in[j^*]} z_i + (m^* - j^*)P + (k-m^*)\pmin}{k}, & i = m^* + 1,\\ 
    \pmin + (c_{m^* +1} - \pmin) (1+{\eta}/{k})^{i-m^*-1}, & i= m^*+2,\dots,i^*,
    \end{cases}\\
    r_i &= \pmin + \frac{\pmax-\pmin}{(1+\gamma/k)^{k-i+1}}, \quad i = i^*+1,\dots,k,
\end{align*}   
\end{subequations}
where $j^* = \left\lceil{\ln\left(\frac{P/\pmin-1}{\gamma - 1}\right)}/{\ln\left(1+\frac{\gamma}{k}\right)} \right\rceil$, $m^* = j^* + \left\lceil \frac{kP/\eta - k\pmin [1 + (\gamma-1)(1+\gamma/k)^{j^*}]/\gamma}{P-\pmin}\right\rceil$, and $i^*$ is the largest index such that $\alpha_{i^*+1}^{(k)}(\phi) = \frac{k r_{i^*+1}}{\sum_{i\in[j^*]}z_i + \sum_{i=j^* + 1}^{i^*} c_i + (k- i^*)\pmin} \le \gamma$.

For interval $i\in[j^*]$, we have $\alpha^{(k)}_i \le \gamma$ since $z_i$ is designed based on Equation~\eqref{eq:pro-beg} in Proposition~\ref{pro:beg}. $j^*$ is determined as the largest index such that $\phi_{j^*} < P \le \pmax$ and thus $j^* \le \xi$. Note that as the prediction $P$ increases, $m^*$ also increases to ensure the consistency. When $P = \pmax$, we have $j^* = \xi$ and $m^*$ is given by
\begin{align*}
    m^* = \xi + \left\lceil \frac{k\pmax/\eta - k\pmin [1 + (\gamma-1)(1+\gamma/k)^{\xi}]/\gamma}{\pmax-\pmin}\right\rceil, 
\end{align*}
which cannot exceed the total number of available units $k$.
Since the target consistency $\eta$ and robustness $\gamma$ are given in Equation~\eqref{eq:robust-consiste-otp}, we have 
\begin{align*}
    &\frac{k\pmax/\eta - k\pmin [1 + (\gamma-1)(1+\gamma/k)^{\xi}]/\gamma}{\pmax-\pmin} \\
    &\quad\quad\quad= k\cdot \frac{\left[1+(\gamma-1)(1+\frac{\gamma}{k})^{\xi}\right]/\gamma + (\theta - 1)(1-\frac{\xi}{k}) - [1 + (\gamma-1)(1+\gamma/k)^{\xi}]/\gamma}{\theta - 1} = k - \xi, 
\end{align*}
and thus the target $\eta$ and $\gamma$ can ensure $m^* \le k$.

For $i = m^* +1$ (that corresponds to the interval $[\phi_{j^{*}+1},\phi_{m^*+1})$), we have
\begin{align*}
   \alpha^{(k)}_{m^*+1} = \frac{k c_{m^*+1}}{\sum_{j\in[j^*]}z_j + (m^* - j^*)P + (k-m^*)\pmin} = \eta.    
\end{align*}

Finally, we have $\alpha^{(k)}_i \le \eta$ for $i = m^*+2,\dots,i^*$ and $\alpha^{(k)}_i \le \gamma$ for $i = i^*+1,\dots,k+1$ by following the same proof as Case II.

In this case, the worst-case ratio $\alpha^{(k)}_{i} \le \eta, i=j^*+1,\dots,i^*$ and $\max_{i\in[k+1]}\alpha^{(k)}_i = \gamma$. Thus, $\ota_{\phi}$ is $\eta$-consistent and $\gamma$-robust.

\subsection{Proof of Theorem~\ref{thm:lb-k-min}}
To show the lower bound, we consider a special family of instances, and show that under the special instances, any $\gamma$-robust online algorithm at least has a consistency $\eta$ lower bounded by $\Lambda(\gamma)$.
\begin{dfn}[reverse $p$-instance]
A reverse $p$-instance $\cali_p$ consists of a sequence of prices that decrease continuously from $p_{\max}$ to $p$ and spike to $p_{\max}$ at the end.   
\end{dfn}

Let $g(p):[\pmin,\pmax]\to \{0,1,\dots,k\}$ denote the cumulative trading decision of an online algorithm when it executes the instance $\cali_p$ before the compulsory trading in the last step. Since online decision is irrevocable, $g(p)$ is non-decreasing as $p$ decreases from $\pmax$ to $\pmin$. In addition, items must be traded at the lowest price $\pmin$. Thus, we must have $g(\pmin) = k$.
Given an online algorithm, let $\cali_{\hat{p}_i}$ denote the first instance, in which the algorithm trades the $i$-th item, i.e., $\hat{p}_i = \sup_{\{p\in[\pmin,\pmax]:g(p)\ge i\}} p$.   

For any $\gamma$-robust online algorithm, $\{\hat{p}_i\}_{i\in[k]}$ must satisfy
\begin{align}
\label{eq:proof-k-min1}
    \frac{\sum_{j\in[i-1]}\hat{p}_j +  (k-i+1) \pmax}{k \hat{p}_i} \le \gamma, i\in[k+1],\quad\quad \hat{p}_i \ge \pmin,  i\in[k],
\end{align}
where $\hat{p}_0:= \pmax$ and $\hat{p}_{k+1}:= \pmin$. 
Based on Equation~\eqref{eq:proof-k-min1}, we have for $i\in[k]$
\begin{align}
  \hat{p}_i \ge \max\left\{\pmax - \pmax\left(1-\frac{1}{\gamma}\right)\left(1+\frac{1}{k\gamma}\right)^{i-1}, \pmin \right\}.
\end{align}

Suppose the prediction is $P=\pmin$. To ensure consistency under the instance $\cali_{\pmin}$, any $\gamma$-robust online algorithm must have
\begin{subequations}
\begin{align}
    \eta \ge \frac{\alg(\cali_{\pmin})}{\opt(\cali_{\pmin})} = \frac{\sum_{i\in[k]} \hat{p}_i}{k \pmin}&\ge \frac{\sum_{i\in[\zeta]} \hat{p}_i + (k-\zeta)\pmin}{k \pmin} \\&\ge \frac{\sum_{i\in[\zeta]} \left[\pmax - \pmax\left(1-\frac{1}{\gamma}\right)\left(1+\frac{1}{k\gamma}\right)^{i-1} \right] +  (k-\zeta)\pmin}{k \pmin}\\
    &= \theta\left[\gamma - (\gamma - 1)\left(1+\frac{1}{\gamma k}\right)^{\zeta}\right] - (\theta - 1)\left(1 - \frac{\zeta}{k}\right),
\end{align}    
\end{subequations}
where $\zeta$ satisfies $\pmax - \pmax\left(1-\frac{1}{\gamma}\right)\left(1+\frac{1}{k\gamma}\right)^{\zeta-1} > \pmin \ge  \pmax - \pmax\left(1-\frac{1}{\gamma}\right)\left(1+\frac{1}{k\gamma}\right)^{\zeta}$, which gives $\zeta = \left\lceil{\ln\left(\frac{\theta-1}{\theta - \theta/\gamma}\right)}/{\ln\left(1+ \frac{1}{\gamma k}\right)} \right\rceil.$

\subsection{Proof of Theorem~\ref{thm:k-min}}

The threshold values $\psi = \{\psi_i\}_{i\in[k]}$ for $k$-min is a non-increasing sequence $\psi_1\ge\psi_2 \ge \dots\ge \psi_k$. 
We also define $\psi_0 = \pmax$ and $\psi_{k+1} = \pmin$ .
The threshold $\psi$ divides the uncertainty range $[\pmin,\pmax]$ into $k+1$ regions.
If $\psi_i = \psi_{i+1} = \dots = \psi_{j} < \psi_{j+1}$, we use $j+1$ to index the interval $[\psi_i, \psi_{j+1})$.
Excluding the compulsory trading at the end, if \ota trades $i$ items, the worst-case price sequence is 
\begin{align*}
    \psi_1, \psi_2, \dots, \psi_{i},\underbrace{\psi_{i+1}+\epsilon,\dots, \psi_{i+1}+\epsilon}_{k}, \underbrace{\pmax,\dots,\pmax}_k,
\end{align*}
where $\epsilon > 0$ and $\epsilon \to 0$. 
The worst-case ratio under this price sequence is
\begin{align}
    \varphi^{(k)}_i:=\varphi^{(k)}_i(\psi) = 
    \frac{\sum_{j\in[i-1]} \psi_{j} + (k-i+1) \pmax}{k\psi_i}.
\end{align}
We start by proving the two following two propositions. 
\begin{pro}
\label{pro:end-min}
For a given $j\in[k-1]$, if the threshold values after $j$ are given by  
\begin{align}
\label{eq:pro-end-min}
 \psi_i = p_{\max} - \frac{p_{\max}-p_{\min}}{(1+ 1/(\gamma k))^{k-i+1}}, i = j+1,\dots, k, 
\end{align}
and $\varphi^{(k)}_{j+1} \le \gamma$, then $\varphi^{(k)}_{i}(\psi) \le \gamma, i = j+2,\dots, k+1.$
\end{pro}
\paragraph{Proof of Proposition~\ref{pro:end-min}.}
Given that $\varphi^{(k)}_{j+1} \le \gamma$, we have
\begin{align}
\label{eq:pro-proof1-min}
    \sum\nolimits_{s\in[j]}\psi_s + (k-j)\pmax \le k\gamma \psi_{j+1}.
\end{align}
We then have
\begin{subequations}
 \begin{align}
    \varphi^{(k)}_{j+2} = \frac{\sum_{s\in[j+1]}\psi_s + (k - j - 1)\pmax}{k \psi_{j+2}} &= \frac{\sum_{s\in[j]}\psi_s + (k - j)\pmax + \psi_{j+1} - \pmax}{k \psi_{j+2}}\\
    \label{eq:pro-proof-app1}
    &\le \frac{k \gamma\psi_{j+1} + \psi_{j+1} - \pmax}{k \psi_{j+2}}\\
    \label{eq:pro-proof-app2}
    &= \gamma \frac{ \pmax + (1+\frac{1}{\gamma k})(\psi_{j+1} - \pmax)}{\psi_{j+2}} = \gamma,
\end{align}   
\end{subequations}
where the inequality~\eqref{eq:pro-proof-app1} holds due to inequality~\eqref{eq:pro-proof1-min} and the last equality~\eqref{eq:pro-proof-app2} is based on the design of the threshold values in equation~\eqref{eq:pro-end-min}.
By following the same step, the worst-case ratios in the following intervals can be shown that $\varphi^{(k)}_i \le \gamma, \forall i = j+3,\dots, k+1$.

\begin{pro}
\label{pro:beg-min}
For a given $j\in[\zeta]$, if the first $j$ threshold values are given by
\begin{align}
\label{eq:pro-beg-min}
 \psi_i = p_{\max} - p_{\max}\left(1  - \frac{1}{\gamma}\right) \left(1 +\frac{1}{\gamma k}\right)^{i-1}, i \in 1,\dots, j, 
\end{align}
then $\varphi^{(k)}_{i}(\psi) \le \gamma, i = 1,\dots, j.$
\end{pro}

\paragraph{Proof of Proposition~\ref{pro:beg-min}.}
When the first $j$ threshold values are designed based on Equation~\eqref{eq:pro-beg-min}, $\forall i\in [j]$, the worst-case ratio of the $i$-th interval of \ota can be shown as
\begin{align}
    \varphi^{(k)}_i = \frac{\sum_{s\in[i-1]}\psi_{s} + (k-i + 1)\pmax}{k \psi_i} = \frac{k\gamma [\pmax - \pmax(1-1/\gamma)(1+1/(k\gamma))^{i-1}]}{k \psi_i} = \gamma.
\end{align}

In the following, we show \ota with the designed threshold $\psi$ can guarantee the target consistency and robustness in the $k$-min search in Cases IV-VI.

\paragraph{Case IV} When $P\in[\tilde{p}_1,\pmax]$, where $\tilde{p}_1 = \pmax - \pmax(1 - 1/\eta)(1 + 1/(\eta k))^{\sigma^* - 1}$ and $\sigma^* \in [k]$ is the largest index such that 
\begin{align}
\label{eq:cond1-min}
    \frac{1 - (1 - 1/\eta)(1+\frac{1}{\eta k})^{\sigma^*}}{1 - (1 - 1/\theta)/(1 + \frac{1}{\gamma k})^{k - \sigma^*}} \le \frac{\gamma}{\eta}. 
\end{align}
In addition, $i^* = \sigma^*$ and $\tilde{p}_1 = c_{i^*}$.
We derive the worst-case ratios $\varphi^{(k)}_i(\psi)$ given the design 
\begin{subequations}
\begin{align*}
c_i &= \pmax - \pmax\left(1  - {1}/{\eta}\right) \left(1 +{1}/{(\eta k)}\right)^{i-1}, \quad i\in[i^*],\\
 r_i &= \pmax - \frac{\pmax-\pmin}{(1+ 1/(\gamma k))^{k-i+1}}, \quad i = i^*+1,\dots,k.
\end{align*}    
\end{subequations}

For $i\in[i^*]$, we have
\begin{align*}
    \varphi^{(k)}_i = \frac{\sum_{j\in[i-1]} c_{j} + (k-i+1) \pmax}{k c_i} = \frac{k \eta c_i }{ k c_i  }= \eta.
\end{align*}

For $i = i^* + 1$, we have
\begin{align*}
    \varphi^{(k)}_{i^*+1} = \frac{\sum_{j\in[i^*]} c_{j} + (k-i^*) \pmax}{k r_{i^*+1}}  &= \frac{k\eta \left[\pmax - \pmax\left(1  - \frac{1}{\eta}\right) \left(1 +\frac{1}{\eta k}\right)^{i^*}\right]}{k  \left[\pmax - \frac{\pmax-\pmin}{(1+ 1/(\gamma k))^{k-i^*}} \right] }\\
    &=  \frac{\eta \left[1 - (1 - 1/\eta)(1+\frac{1}{\eta k})^{i^*}\right]}{1 - (1 - 1/\theta)/(1 + \frac{1}{\gamma k})^{k - i^*}} \le \gamma,
\end{align*}
where the last inequality is due to Equation~\eqref{eq:cond1-min}.

For $i = i^*+2,\dots, k+1$, based on Proposition~\ref{pro:end-min} and $\varphi^{(k)}_{i^*+1} \le \gamma$, we have $\varphi^{(k)}_i \le \gamma,$ for all $i = i^*+2,\dots, k+1$.

Summarizing above results, in this case, we have $\varphi^{(k)}_i \le \eta, i\in [i^*]$ and thus $\ota_{\psi}$ is $\eta$-consistent. The worst-case competitive ratio is $\max_{i\in[k+1]} \varphi^{(k)}_i =\gamma$ and thus $\ota_{\psi}$ is $\gamma$-robust.

\paragraph{Case V} When the prediction $P\in [\tilde{p}_2,\tilde{p}_1)$, where $\tilde{p}_2 = \min\{\tilde{p}_1, \pmax/ \gamma\}$. This case exists only when $\tilde{p}_1 > \pmax/\gamma $.
In the following, we derive the worst-case ratios $\varphi^{(k)}_i$ given the threshold design
\begin{subequations}
\begin{align*}
c_i&= \begin{cases}
P, & i = 1,\dots,m^*,\\
 \frac{m^* P + (k-m^*)\pmax}{\eta k}, &  i= m^* + 1,\\
\pmax - (\pmax - c_{m^* +1}) \left(1 + {1}/{(\eta k)} \right)^{i-m^*-1}, & i= m^*+2,\dots,i^*,
\end{cases}\\
r_i &= \pmax - \frac{\pmax-\pmin}{(1+ 1/(\gamma k))^{k-i+1}}, \quad i = i^*+1,\dots,k,
\end{align*}    
\end{subequations}
where $m^* = \left\lceil k\frac{\pmax- \eta P }{\pmax-P}\right\rceil$ and $i^*$ is the largest index that ensures
$\varphi^{(k)}_{i^*+1}(\psi)\le \gamma.$

For $i = 1$, $\varphi^{(k)}_1 = (k \pmax)/k P   \le \pmax/\tilde{p}_1 \le \gamma$. 

For $i = m^*+1$, 
\begin{align*}
    \varphi^{(k)}_{m^*+1} = \frac{m^* P + (k-m^*)\pmax}{k c_{m^*+1}} = \eta.
\end{align*}

For $i= m^* + 2, \dots, i^*$, we have
\begin{align*}
    \varphi_{i} = \frac{m^* P + \sum_{j=m^*+1}^{i-1} c_j + (k-i + 1)\pmax}{k c_{i}}= \frac{k \eta \left[\pmax - (\pmax - c_{m^* +1}) \left(1 + \frac{1}{\eta k} \right)^{i-m^*-1}\right]}{k c_{i}} = \eta.
\end{align*}

For $i = i^* + 1, \dots, k+1$, based on $\varphi^{(k)}_{i^*+1} \le \gamma$ and Proposition~\ref{pro:end-min}, we have $\varphi^{(k)}_i \le \gamma$.

Based on above results, when prediction $P\in [\tilde{p}_2,\tilde{p}_1)$, the consistency of $\ota_{\psi}$ is $\eta$. Since $\max_{i\in[k+1]} \varphi^{(k)}_i \le \gamma$, the robustness is $\gamma$.

\paragraph{Case VI} When $P\in [\pmin,\tilde{p}_2)$. 
The worst-case ratios $\varphi^{(k)}_i$ can be shown as follows when thresholds are 
\begin{subequations}
\begin{align*}
z_i & = \pmax - \pmax\left(1 - {1}/{\gamma}  \right) \left(1+{1}/{(\gamma k)}\right)^{i-1},\quad i\in[j^*],\\
c_i&= \begin{cases}
P, & i = j^*+1,\dots,m^*,\\
\frac{\sum_{i\in[j^*]} z_i + (m^* - j^*)P + (k-m^*)\pmax}{\eta k },& i = m^* + 1,\\ 
\pmax - (\pmax - c_{m^* +1}) \left(1 + {1}/{(\eta k)}\right)^{i-1}, & i= m^*+2,\dots,i^*,
\end{cases}\\
r_i &= \pmax - \frac{\pmax-\pmin}{(1+ 1/(\gamma k))^{k-i+1}},\quad i = i^*+1,\dots,k,
\end{align*}    
\end{subequations}
where $j^* = \left\lceil{\ln\left(\frac{1 - P/\pmax}{1 - 1/\gamma }\right)}/{\ln\left(1+{1}/{(\gamma k)}\right)} \right\rceil$, and $m^* = j^* + \left\lceil \frac{k\pmax  [1 - (1 - 1/\gamma)(1+1/(\gamma k))^{j^*}] - kP\eta}{\pmax - P}\right\rceil$, and $i^*$ is the largest index such that $\varphi^{(k)}_{i^*+1}(\psi) \le \gamma$.

For interval $i\in[j^*]$, we have $\varphi^{(k)}_i \le \gamma$ since $z_i$ is designed based on Equation~\eqref{eq:pro-beg-min} in Proposition~\ref{pro:beg-min}. $j^*$ is determined as the largest index such that $\psi_{j^*} > P \ge \pmin$.

For interval $i = m^* +1$ (i.e., $[\psi_{j^{*}+1},\psi_{m^*+1})$), we have
\begin{align*}
   \varphi^{(k)}_{m^*+1} = \frac{\sum_{j\in[j^*]}z_j + (m^* - j^*)P + (k-m^*)\pmax}{k c_{m^*+1}} = \eta.    
\end{align*}

Finally, we have $\varphi^{(k)}_i \le \eta$ for interval $i = m^*+2,\dots,i^*$ and $\varphi^{(k)}_i \le \gamma$ for interval $i = i^*+1,\dots,k+1$ by following the same proof as Case V.

In this case, 
the worst-case ratio $\varphi^{(k)}_{i} \le \eta, i=j^*+1,\dots,i^*$ and $\max_{i\in[k+1]}\varphi^{(k)}_i = \gamma$. Thus, $\ota_{\psi}$ is $\eta$-consistent and $\gamma$-robust.

\section{Proofs on online search with inventory dynamics}
\label{app:inventory}

\subsection{Proof of Lemma~\ref{lem:feasible}}
First, we can observe the inventory dynamics is enforced in Line~\ref{alg-line:dynamics} of Algorithm~\ref{alg:osid}. Since the online decision is $x_t = \max\left\{\sum_{j\in[i]} x_{t}^{(n,j)}, d_t - s_{t-1}\right\}$ (Line~\ref{alg-line:online-sol}), we have $x_t \ge d_t - s_{t-1}$ and thus $s_t \ge 0, \forall t\in[T]$. Therefore, we only need to check the capacity constraint of the inventory. 
For any $t$ in a busy period of the $n$-th interval with $i$ parallel online search problems, observe 
\begin{align*}
   s_t &=  \sum\nolimits_{j=1}^{i} (m^{(n,j)}_t - 1) - \sum\nolimits_{j = 2}^{i} k^{(n,j)} \\
   &\le \sum\nolimits_{j=1}^{i} k^{(n,j)} - \sum\nolimits_{j = 2}^{i} k^{(n,j)} = k^{(n,1)} = B, 
\end{align*}
where $m^{(n,j)}_t - 1$ denotes the number of purchased items of the $j$-th online search problem up to time $t$, and $m^{(n,j)}_t - 1 \le k^{(n,j)}$ in the online search problem.
Thus, $\sum\nolimits_{j=1}^{i} (m^{(n,j)} - 1)$ and $\sum\nolimits_{j = 2}^{i} k^{(n,j)}$ represent the total number of purchased items and the total demand from the start of the busy period to slot $t$, respectively. Thus, the inventory level never violates the capacity constraint.

\subsection{Proof of Lemma~\ref{lem:ub-alg-oi}}
The total cost of the online algorithm can be upper bounded by the costs of purchasing items during the busy period and idle period.
In the busy period of interval $n$, there are a total of $i_n$ virtual search problems and the $i$-th problem buys $m^{(n,i)} -1$ items. Since each virtual search problem buys items based on the \ota, the price of $j$-th purchased item of problem $i$ in interval $n$ is at most $\psi^{(n,i)}_j$. Thus, the cost of busy period is upper bounded by  $\sum_{n\in[N]}\sum_{i\in[i_n]}\sum_{j\in [m^{(n,i)}-1]} \psi^{(n,i)}_j$. 
In addition, the price of purchasing items in the busy period is upper bounded by $\pmax$. Thus,
\begin{subequations}
 \begin{align}
{\alg}(\cali) &\le \sum_{n\in[N]}\sum_{i\in[i_n]}\sum_{j\in [m^{(n,i)}-1]} \psi^{(n,i)}_j + \biggl[\sum_{t\in[T]} d_t + s_T - \sum_{n\in[N]}\sum_{i\in[i_n]} (m^{(n,i)} - 1) \biggr] p_{\max}\\
&= \sum_{n\in[N]}\sum_{i\in[i_n]} \biggl[ \sum_{j\in [m^{(n,i)}-1]} \psi^{(n,i)}_j + (k^{(n,i)} - m^{(n,i)} + 1) p_{\max} \biggr] + \left[R + s_T\right] p_{\max},
\end{align}   
\end{subequations}
where $R := \sum_{t\in[T]} d_t - \sum_{n\in[N]}\sum_{i\in[i_n]} k^{(n,i)} = \sum_{n\in[N]}\sum_{t\in\calt_n^0} d_t - NB$.

\subsection{Proof of Lemma~\ref{lem:lb-opt-oi}}
Let $F(\hat{s}_n, y_{n})$ denote the cost of the offline optimum in the busy period of the interval $n$, where $\hat{s}_n$ is the initial inventory in the beginning of the interval $n$ and $y_n$ is the number of purchased items in this busy period. Let $F^*(\hat{s}_n, y_{n})$ denote the minimum cost of purchasing $y_{n}$ items with initial inventory $\hat{s}_n$ in the busy period of interval $n$ without considering the price and demand of other intervals. Then we can have 
\begin{align}\label{eq:lem4}
    F_n(\hat{s}_n, y_{n}) \ge F^*_n(\hat{s}_n, y_{n}) \ge F^*_n(0, y_{n}), \forall n\in[N],
\end{align}
where the first inequality is by definition and the second inequality is because a lower initial inventory gives more flexibility to minimize the cost of purchasing the same amount of items.

Further, we can have, for all $n\in[N]$,
\begin{align}\label{eq:lem1}
   F^*_n\left(0, \sum\nolimits_{i\in[i_n]}k^{(n,i)}\right) - F^*_n(0, y_n) \le \left(\sum\nolimits_{i\in[i_n]}k^{(n,i)} - y_n\right) \frac{p_{\max}}{\gamma}.
\end{align}
Note that $\sum\nolimits_{i\in[i_n]}k^{(n,i)}$ is the sum of inventory capacity and the total demand in the busy period. Thus, $y_n \le \sum\nolimits_{i\in[i_n]}k^{(n,i)}$.
In addition, there must exist prices no larger than $\frac{p_{\max}}{\gamma}$ in the busy period. Therefore, above inequality holds since the price for purchasing any additional items in the busy period by the cost minimum algorithm will be no larger than $\frac{p_{\max}}{\gamma}$. In addition, we have
\begin{align}\label{eq:lem2}
    F^*_n\left(0, \sum\nolimits_{i\in[i_n]}k^{(n,i)}\right) \ge \sum\nolimits_{i\in [i_n]} \psi^{(n,i)}_{m^{(n,i)}} k^{(n,i)}, 
\end{align}
because $\psi^{(n,i)}_{m^{(n,i)}}$ is the minimum price of the $i$-th virtual search problem of interval $n$ in its lifecycle.
Combining Equations~\eqref{eq:lem4}-~\eqref{eq:lem2} gives
\begin{align}\label{eq:lem5}
    F_n(\hat{s}_n, y_{n}) \ge \sum\nolimits_{i\in [i_n]} \psi^{(n,i)}_{m^{(n,i)}} k^{(n,i)} - \left[\sum\nolimits_{i\in[i_n]}k^{(n,i)} - y_n\right] \frac{\pmax}{\gamma}.
\end{align}

The offline optimal cost can be lower bounded by
\begin{subequations}
\begin{align}
    {\opt}(\cali) &\ge \sum\nolimits_{n\in[N]} F_n(\hat{s}_n, y_{n}) + \left[\sum\nolimits_{t\in[T]} d_t - \sum\nolimits_{n\in[N]} y_n\right] \frac{\pmax}{\gamma} \\
    &\ge \sum\nolimits_{n\in[N]}\sum\nolimits_{i\in[i_n]} 
\psi^{(n,i)}_{m^{(n,i)}} k^{(n,i)} + \frac{R \cdot p_{\max}}{\gamma},
\end{align}   
\end{subequations}
where the first inequality holds since the minimum price during idle period is $\frac{\pmax}{\gamma}$, and the second inequality is obtained by substituting inequality~\eqref{eq:lem5}.

\section{Continuous version of online search with inventory dynamics}
\label{app:osid}

We present algorithms and results for the continuous version of the online search with inventory dynamics (\osid).
In this problem, integral constraints on the purchasing decisions and inventory levels are relaxed, i.e., the constraint~\eqref{eq:osid4} is relaxed to $x_t \ge 0, s_t \ge 0, \forall t\in[T]$.
We show that the learning-augmented algorithm for \osid (i.e., Algorithm~\ref{alg:osid}) can be extended to solve the continuous \osid and achieve the Pareto-optimal trade-off between consistency and robustness.
In Section~\ref{sec:numerics}, we further apply the extended algorithm to solve the storage-assisted energy procurement problem for data centers in the electricity markets.  
\paragraph{Learning-augmented algorithm for continuous \osid}
We use a modified Algorithm~\ref{alg:osid} to solve the continuous \osid. Specifically, we choose a large integer $K$ as an additional input.
In this algorithm, for each created virtual search problem $(n,i)$ with prediction $P^{(n,i)}$, we treat it as a $(K,P^{(n,i)})$-search problem, and use the threshold values $\psi(K,P^{(n,i)})$ designed in Equations~\eqref{eq:psi-I}-\eqref{eq:psi-III} to solve each virtual problem in line~\ref{alg:virtual-start} to line~\ref{alg:virtual-end} of Algorithm~\ref{alg:osid}. Then we change the online decisions in line~\ref{alg-line:online-sol} by scaling them back from the virtual problems, i.e.,
\begin{align}
    x_t = \max\left\{\sum_{j\in[i]} \frac{x_{t}^{(n,j)}\cdot k^{(n,j)} }{K}, d_t - s_{t-1}\right\},
\end{align}
where $k^{(n,i)}$ is the capacity of the virtual search problem.

\paragraph{Consistency-robustness results}
We first note that $K$-min search ($K\to\infty$) is a special case of the continuous \osid. 
Based on Theorem~\ref{thm:lb-k-min}, we can have the following lower bound result. 
\begin{cor}
    For continuous \osid with predictions, any $\gamma$-robust deterministic learning-augmented algorithm must have a consistency lower bounded by $\Lambda(\gamma) = \gamma - (\theta - 1) \left[1 - \gamma\ln\left(\frac{\theta - 1}{\theta - \theta/\gamma} \right) \right]$.
\end{cor}

Based on this lower bound, for a given $\lambda \in [0,1]$, we set the target consistency and robustness as 
\begin{align}
\label{eq:consist-robust-cmin}
\gamma^{(\infty)}(\lambda) = \varphi^{(\infty)}_* + (1-\lambda)(\theta - \varphi^{(\infty)}_*), \quad \eta^{(\infty)}(\lambda) = \Lambda(\gamma^{(\infty)}(\lambda)),
\end{align}
where $\varphi^{(\infty)}_*$ is the optimal competitive ratio of $K$-min search (as $K\to\infty$), and is the solution of $\frac{1 - 1/\theta}{1 - 1/\varphi} = \exp\left(\frac{1}{\varphi}\right)$.

Based on upper bound results in Theorem~\ref{thm:k-min}, we further have the consistency-robustness result for the modified Algorithm~\ref{alg:osid}. 
\begin{cor}
\label{thm:c-osid}
Given $\lambda \in[0,1]$, the modified Algorithm~\ref{alg:osid} (with $K\to\infty$) is $\eta^{(\infty)}(\lambda)$-consistent and $\gamma^{(\infty)}(\lambda)$-robust for the learning-augmented continuous \osid when ${\psi}_{K,P}(\cdot)$ are given by Equations~\eqref{eq:psi-I}-\eqref{eq:psi-III}. The modified Algorithm~\ref{alg:osid} is Pareto-optimal.
\end{cor}

As a remark, the existing work~\cite{yang2020online} studies this continuous \osid and designs an online algorithm that achieves the optimal competitive ratio. 
Algorithm~\ref{alg:osid} can be considered as the learning-augmented extension for the worst-case optimized algorithm. In addition, Algorithm~\ref{alg:osid} can also achieve the optimal competitive ratio for \osid when we set $\lambda = 1$ to ignore the predictions.
\section{$k$-search experimental results in general financial markets}
\label{sec:bitcoin}
\begin{figure*}[t]
    \begin{minipage}[b]{.99\textwidth}
    \centering
    \vspace{0mm}
    \subfigure[Prediction quality]{\label{fig-max-boxplot}\includegraphics[width=.3\textwidth]{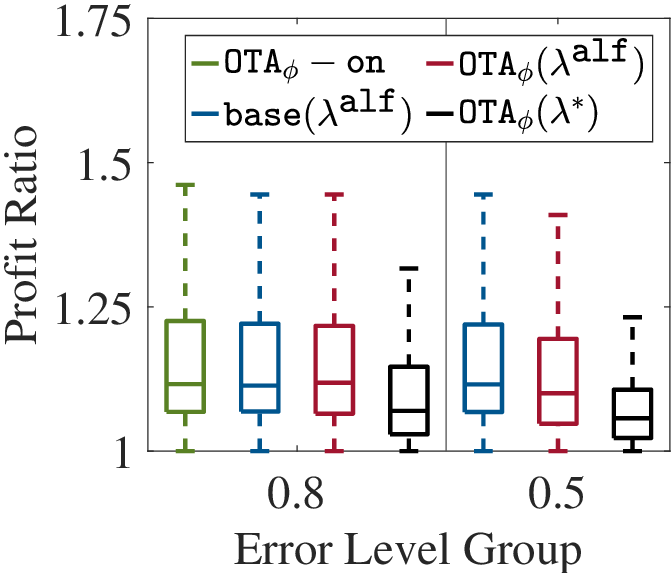}}
    \vspace{-2mm}
    \subfigure[Instance hardness]{\label{fig-max-boxplot-group}\includegraphics[width=.3\textwidth]{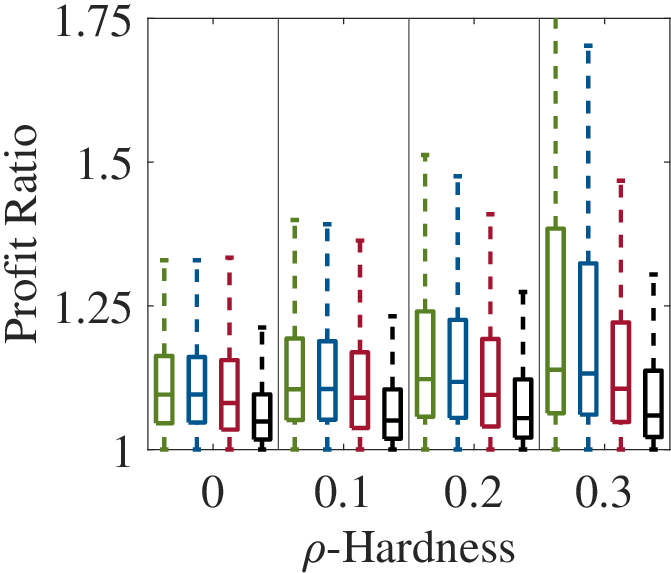}\vspace{-4mm}}
    \caption{Comparing algorithms for $k$-max search.}
    \label{fig-max-all}
    \end{minipage}\\
    \begin{minipage}[b]{.99\textwidth}\vspace{0mm}
    \centering
    \subfigure[Prediction quality]{\label{fig-min-boxplot}\includegraphics[width=.3\textwidth]{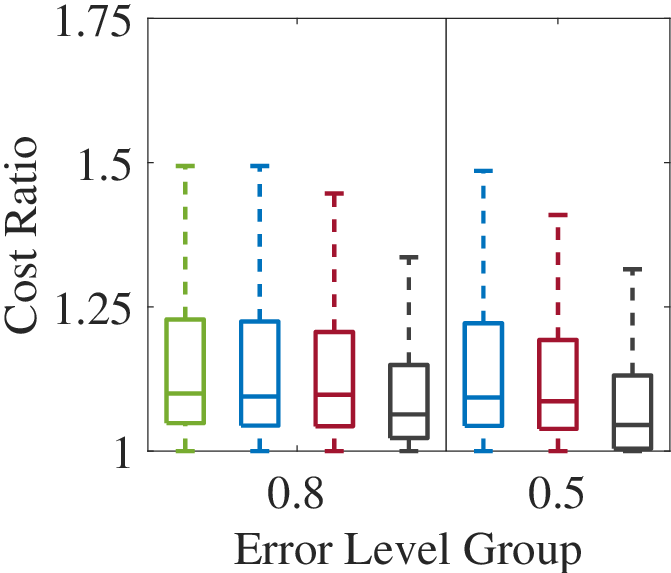}\vspace{-4mm}}
    \vspace{-2mm}
    \subfigure[Instance hardness]{\label{fig-min-boxplot-group}\includegraphics[width=.3\textwidth]{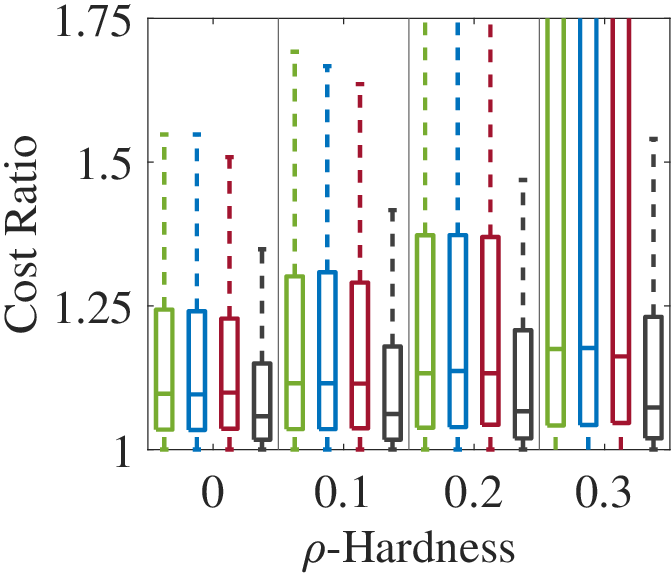}\vspace{-4mm}}
    \caption{Comparing algorithms for $k$-min search.}
    \label{fig-min-all}
    \end{minipage}
\end{figure*}

\begin{figure*}[t]
   \begin{minipage}[b]{.45\textwidth}
   \centering
    \includegraphics[width=.7\textwidth]{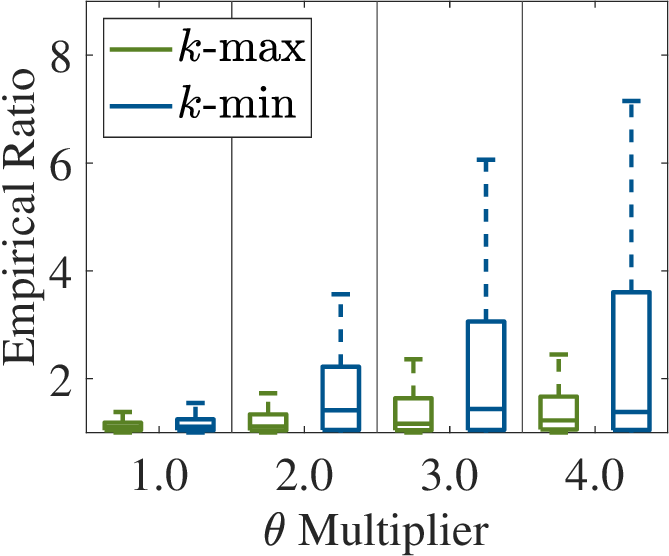}
    \vspace{-4mm}
    \caption{Impact of price uncertainty}
    \label{fig-compare-theta}
    \end{minipage}\hspace{0mm}
    \quad\quad\quad\quad
    \begin{minipage}[b]{.45\textwidth}
    \centering
    \includegraphics[width=.7\textwidth]{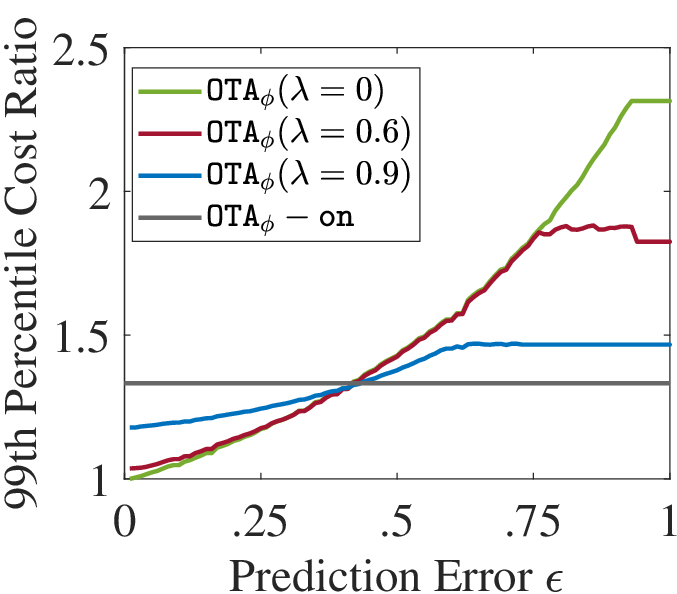}
    \vspace{-4mm}
    \caption{Impact of prediction quality}
    \label{fig-error-parameter}
    \end{minipage}\hspace{0mm}
\end{figure*}

We present our main results for the application of trading in electricity markets, but our algorithms are applicable to general online search problems.  In the additional experiments in this section, we evaluate our algorithms for learning-augmented $k$-search in a general financial market.  We use a 5-year history of Bitcoin (BTC) prices collected from the Gemini Exchange \cite{gemini} in this application.

\subsection{Experimental setup}  We consider the problem of buying or selling BTC over 3-week trading periods. Prices are observed every $10$ minutes such that $T = 3024$.  We set the units available to be traded to $k = 100$.  The price limits $\pmax$ and $\pmin$ are conservatively set according to the maximum and minimum over the entire 5-year period.

To generate a prediction $P$, we use the observed maximum (or minimum) exchange rate of the previous instance.  To evaluate the impact of prediction quality, we adjust the prediction error level with a scalar value between 0 and 1, where 0 indicates perfect predictions and 1 indicates non-adjusted predictions. 
Although BTC has experienced drastic price fluctuations, these occurrences are still rare.  Therefore, to evaluate the performance in worst-case settings, we define an instance $\rho-$hard by giving it a fixed probability $\rho$ of observing the worst-case price in the last time slot.  
\subsection{Comparison algorithms}
We compare the following four algorithms.
\begin{itemize}
    \item[$\triangleright$] (\otpalgwc) The worst-case optimized online algorithm that does not take into account predictions, but guarantees optimal competitive ratios.
    \item[$\triangleright$] (\otpalgoff) Our proposed algorithm with the best possible hyper-parameter $\la^*$ chosen offline. This algorithm is not practical since it is fed with the optimal parameter, but it represents the best improvement from available predictions.
    \item[$\triangleright$](\otpalgalf) Our proposed algorithm that uses an online learning algorithm (i.e., adversarial Lipschitz algorithm~\cite{maillard2010}) to adaptively select $\la$.
    \item[$\triangleright$](\naivebaseline) The baseline algorithm that is introduced in Section~\ref{sec:learning-augmented-alg} and uses the same online learning approach to select $\lambda$.
\end{itemize}

{
\subsection{Experimental results}
We compare the empirical competitive ratios of four algorithms for $k$-max and $k$-min search problems in Figures~\ref{fig-max-all} and~\ref{fig-min-all}, respectively.
Figures~\ref{fig-max-boxplot} and~\ref{fig-min-boxplot} show \otpalgalf can make the best use of prediction among all algorithms and benefit most as the prediction quality improves (from $0.8$ to $0.5$).
In addition to incorporating good predictions, \otpalgalf also maintains the best robustness, which degrades slowly when the instances become harder, as shown in Figures~\ref{fig-max-boxplot-group} and~\ref{fig-min-boxplot-group}.
Thus, the experiments show the good potential of \otpalgalf to achieve the best of both worlds.
Since $k$-max and $k$-min are known to have different worst-case performances~\cite{lorenz2009optimal}, we further compare \otpalgalf for $k$-max and $k$-min. The empirical performance of $k$-min is generally worse than $k$-max, which is most apparent for larger values of uncertainty parameter $\theta$ in Figure \ref{fig-compare-theta}. This is consistent with the worse consistency-robustness trade-off of $k$-min as shown in Figure~\ref{fig:pareto-boundary}.
In Figure~\ref{fig-error-parameter}, we further investigate the impact of prediction quality on the worst-case performance of our algorithms. Specifically, we consider 100 instances of $k$-max search while adding prediction errors $\epsilon$ (normalized by $\pmax-\pmin$), and report the $99$-th percentile of the empirical ratios. We can observe that even when the prediction is almost incorrect by $(\pmax-\pmin)/2$, our algorithms can still outperform the worst-case optimized algorithm \otpalgwc.
}

\end{document}